\pdfoutput=1
\documentclass{article}
\PassOptionsToPackage{numbers,square}{natbib}
\usepackage[final,main]{neurips_2026}
\usepackage[utf8]{inputenc}
\usepackage[T1]{fontenc}
\usepackage{microtype}
\usepackage[english]{babel}
\usepackage[dvipsnames,table]{xcolor}
\usepackage{graphicx}
\usepackage{wrapfig}
\usepackage{booktabs,multirow,multicol}
\usepackage{amsmath,amsfonts}
\usepackage{hyperref}
\usepackage[nameinlink,capitalize]{cleveref}

\usepackage{amsthm}

\usepackage{centernot}
\usepackage{nicefrac}       %
\usepackage{mathtools}
\usepackage{amsbsy}
\usepackage{amstext}
\usepackage{thmtools}
\usepackage{thm-restate}

\begingroup
    \makeatletter
    \@for\theoremstyle:=definition,remark,plain\do{%
        \expandafter\g@addto@macro\csname th@\theoremstyle\endcsname{%
            \addtolength\thm@preskip\parskip
            }%
        }
\endgroup

\crefname{lemma}{lemma}{lemmas}
\Crefname{lemma}{Lemma}{Lemmas}
\crefname{thm}{theorem}{theorems}
\Crefname{thm}{Theorem}{Theorems}
\crefname{prop}{proposition}{propositions}
\Crefname{prop}{Proposition}{Propositions}
\crefname{assumption}{assumption}{assumptions}
\crefname{assumption}{Assumption}{Assumptions}

\renewcommand{\mid}{~\vert~}

\usepackage{booktabs,arydshln}
\makeatletter
\def\adl@drawiv#1#2#3{%
        \hskip.5\tabcolsep
        \xleaders#3{#2.5\@tempdimb #1{1}#2.5\@tempdimb}%
                #2\z@ plus1fil minus1fil\relax
        \hskip.5\tabcolsep}
\newcommand{\cdashlinelr}[1]{%
  \noalign{\vskip\aboverulesep
           \global\let\@dashdrawstore\adl@draw
           \global\let\adl@draw\adl@drawiv}
  \cdashline{#1}
  \noalign{\global\let\adl@draw\@dashdrawstore
           \vskip\belowrulesep}}
\makeatother

\renewcommand{\epsilon}{\varepsilon}

\declaretheorem[style=plain,name=Theorem]{theorem}
\declaretheorem[style=plain,sibling=theorem,name=Lemma]{lemma}
\declaretheorem[style=plain,sibling=theorem,name=Proposition]{proposition}

\declaretheorem[style=definition,sibling=theorem,name=Example]{example}

\newenvironment{example*}
 {\pushQED{\qed}\example}
 {\popQED\endexample}
\numberwithin{equation}{section}

\DeclarePairedDelimiterX\Set[1]{\lbrace}{\rbrace}%
{  #1 }

\usepackage{thm-restate}
\usepackage{fontawesome}
\usepackage{cleveref}
\usepackage{etoc}
\usepackage{amssymb}
\usepackage{placeins}
\usepackage{tabularx}
\usepackage[framemethod=tikz]{mdframed}

\crefname{appsec}{Appendix}{Appendices}
\Crefname{appsec}{Appendix}{Appendices}

\crefformat{appsec}{#2Appendix~#1#3}
\Crefformat{appsec}{#2Appendix~#1#3}

\crefformat{equation}{(#2#1#3)}
\crefformat{figure}{Figure~#2#1#3}
\crefname{example}{Example}{Examples}

\newtheorem{corollary}{Corollary}

\usepackage{enumitem} %
\usepackage[separate-uncertainty=true,multi-part-units=single]{siunitx} %

\declaretheoremstyle[
spacebelow=\parsep,
    spaceabove=\parsep,
  mdframed={
    backgroundcolor=gray!10!white,     %
    hidealllines=true,
    innertopmargin=8pt,
    innerbottommargin=4pt,
    skipabove=8pt,
    skipbelow=10pt,
    nobreak=true
}
]{grayboxed}

\definecolor{takeawaybackground}{HTML}{F2F6FC}
\definecolor{takeawayborder}{HTML}{6B8CC4}
\newcommand{\takeaway}[1]{%
\begin{mdframed}[
    backgroundcolor=takeawaybackground,
    linecolor=takeawayborder,
    linewidth=0.5pt,
    roundcorner=3pt,
    innerleftmargin=8pt,
    innerrightmargin=8pt,
    innertopmargin=6pt,
    innerbottommargin=6pt,
    skipabove=8pt,
    skipbelow=8pt,
    nobreak=true
]
\noindent\textbf{Takeaway.} #1
\end{mdframed}%
}
\crefname{gassumption}{Assumption}{Assumptions}

\usepackage{xcolor}

\definecolor{WowColor}{rgb}{.75,0,.75}
\definecolor{SubtleColor}{rgb}{0,0,.50}

\newcounter{margincounter}

\usepackage{makecell}

\definecolor{mydarkblue}{rgb}{0,0.08,0.45}
\hypersetup{ %
    pdftitle={},
    pdfauthor={},
    pdfsubject={},
    pdfkeywords={},
    pdfborder=0 0 0,
    pdfpagemode=UseNone,
    colorlinks=true,
    linkcolor=mydarkblue,
    citecolor=mydarkblue,
    filecolor=mydarkblue,
    urlcolor=mydarkblue,
    pdfview=FitH
}

\usepackage{multirow}
\usepackage{tikz}
\usetikzlibrary{positioning}
\usetikzlibrary{arrows, automata}
\usetikzlibrary{patterns}

\def\showauthornotes{1}
\ifnum\showauthornotes=1
\newcommand{\Authornote}[2]{{\sf\small\color{blue}{[#1: #2]}}}
\else
\newcommand{\Authornote}[2]{}
\fi

\title{Tight PAC-Bayes Generalisation Guarantees\\for Large Language Model Judges}

\author{%
  Tom A. Lamb\\
  University of Oxford
  \And
  Philip H. S. Torr\\
  University of Oxford
  \And
  Tim G. J. Rudner\\
  University of Toronto,\\
  Vector Institute, Vijil
}

\date{}

\begin{document}

\maketitle
\nointerlineskip
\vbox to 0pt{%
  \vss
  \hbox to \textwidth{\hfil 
  
  {\normalfont
  \href{https://tomalamb.github.io/pac-bayes-judges/}{\faGlobe\enspace Website} \quad
  \href{https://github.com/tomalamb/generalisation-guarantees-llm-judges}{\faGithub\enspace Code}
  }
  \hfil}%
  \vskip 0.6em
}

\renewcommand{\thefootnote}{}
\footnotetext{Correspondence to \href{mailto:thomas.lamb@eng.ox.ac.uk}{\texttt{{thomas.lamb@eng.ox.ac.uk}}} and \href{mailto:tim.rudner@utoronto.ca}{\texttt{tim.rudner@utoronto.ca}}.}
\renewcommand{\thefootnote}{\arabic{footnote}}

\begin{abstract}
\noindent Large language model (LLM) judges are increasingly used to monitor the outputs of LLM-based AI agents, yet formal guarantees on the reliability of these judges remain largely absent, limiting confidence in their use for safety-critical oversight. We consider certifying such a judge and define its risk over safety classifications of prompt--response pairs drawn from a deployed generator. We apply a PAC-Bayes framework to this setting, refining existing compression-based methodology with the use of tighter bounds and more efficient coding. We show that standard parameter-efficient fine-tuning can yield non-vacuous and even tight certificates for safety judges, for both classification and Brier risk, without holding out data for validation. We introduce LoRA-GT, which shares a single quantisation scale across layers and achieves the tightest certificates with minimal degradation in predictive performance. Across five guardrail models and four benchmarks, we find that certificate tightness is governed more by adaptation complexity than by base model scale. Finally, we show how far quantisation can be pushed before the resulting increase in risk outweighs the bound tightening gained from shorter descriptions. Together, these results provide a practical route toward certifiable scalable oversight and offer insight into how the complexity of safety adaptations shapes generalisation.
\end{abstract}

\etocsettocdepth.toc{none}

\section{Introduction}

Large language models (LLMs) are increasingly deployed in generative workflows \citep{brown2020language} and agentic systems \citep{luo2025large}, including in safety-critical domains, yet remain vulnerable to adversarial attacks, misuse, and hallucinations \citep{wang2023decodingtrust, zou2023universal} and so require oversight. Since human supervision does not scale, \textit{scalable oversight} \citep{bowman2022measuring}, in which LLMs act as automated judges, is becoming increasingly common for monitoring deployed models \citep{inan2023llama}. Guardrail models such as Llama Guard \citep{chi2024llama} and Qwen3Guard \citep{zhao2025qwen3guard} already \mbox{make safety decisions at scale, such as whether to block, escalate, or allow a model output.}

For these decisions to be reliable, judges must generalise beyond their evaluation samples, while thresholded safety decisions additionally require well-calibrated uncertainty estimates \citep{gneiting2007strictly, geifman2017selective, guo2017calibration}. However, judges are increasingly used in safety-critical settings without formal guarantees on their performance or predictive uncertainty, while what determines their generalisation remains poorly understood. Validation-based bounds can provide guarantees for a fixed predictor using data held out from training, but can be costly when labelled safety data are limited, and depend only on held-out performance and sample size. They therefore provide little insight into which properties of the learned model and learning process are associated with strong generalisation.

Guardrail models such as Llama Guard \citep{inan2023llama} are already capable safety judges, so adaptation to a new safety taxonomy or set of guidelines can be viewed as a simple refinement of their existing safety capabilities. This raises the question of whether simple adaptations can achieve both strong predictive performance and tight generalisation guarantees without holding out limited safety data. PAC-Bayes theory provides a natural route to investigating this question, and to understanding the role of adaptation simplicity in generalisation, by relating empirical performance to model complexity \citep{alquier2024user,catoni2007pac}. Compression provides a practical way to control this complexity through the description length of a learned model, and recent compression-based analyses have obtained non-vacuous guarantees at the scale of generative LLMs \citep{lotfi2024non, lotfi2024unlocking}. These works, however, focus on a model's own generative risk and often use specialised compressible parameterisations for large-scale pre-training. This motivates our central question: \textit{can we obtain practical, non-vacuous guarantees for safety judges using simple, standard adaptation pipelines and the same data for both training and certification?}

\begin{figure}[t]
    \centering

    \includegraphics[width=1.0\linewidth]{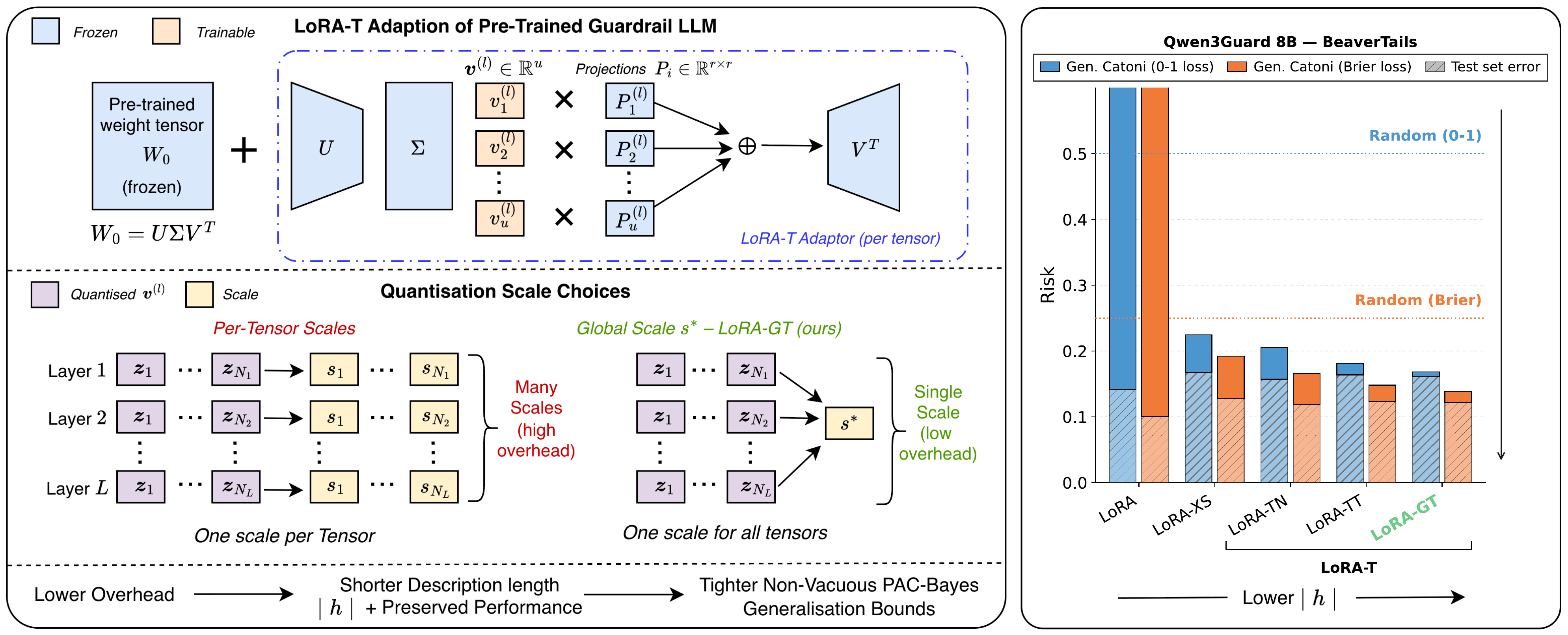}
\vspace{-5pt}
\caption{
\textbf{Compression yields tight, non-vacuous PAC-Bayes guarantees for safety oversight models.}
\textbf{Left:} LoRA-T with per-tensor versus shared global scaling (LoRA-GT), which removes the per-layer quantisation-scale metadata that dominates PEFT adaptor description lengths under aggressive compression.
\textbf{Right:} certified bounds and test risk on BeaverTails under $0$--$1$ and Brier losses, where LoRA-GT achieves the tightest certificates among the shown adaptors while retaining competitive test risk.
}
\label{fig:pull}
\end{figure}

We develop a PAC-Bayes framework for certifying a judge’s risk over prompt--response pairs drawn from the distribution induced by a fixed generator. We use parameter-efficient fine-tuning (PEFT) \citep{hu2022lora} to obtain low-dimensional adaptations that can be readily quantised and compressed. The framework requires minimal changes to standard safety-fine-tuning pipelines, as the judge is trained conventionally before its adaptor is quantised and compressed for certification. \Cref{fig:pull} shows this approach, with LoRA-GT removing the per-layer quantisation-scale metadata that dominates adaptor description length under aggressive compression, yielding the tightest certificate while retaining competitive predictive performance.

Our contributions are the following:\vspace*{-5pt}
\begin{enumerate}[leftmargin=2em]
\item
We formalise PAC-Bayes certification for judges evaluating prompt--response pairs drawn from a fixed deployed generator, and certify both 0--1 classification risk and Brier risk, to obtain certificates for both predictive accuracy and calibration.

\item
We introduce LoRA-GT, which shares a single quantisation scale across layers and yields the tightest certificates in all 40 model--dataset--loss cases. We also tighten the certification methodology with a generalised Catoni bound and Elias $\delta$ coding, and shorten adaptor descriptions using Brotli compression, further improving bound tightness over prior methodology.

\item
Across five guardrail models and four benchmarks, we show that tailored PEFT can achieve strong predictive performance and non-vacuous, often tight certificates, even in small-data regimes without requiring a separate held-out validation set, making these guarantees practical when labelled safety data are limited.
Finally, we show that certificate tightness depends more on adaptation complexity than on base model scale, suggesting that simple adaptations to sufficiently capable base models are sufficient for reliable safety judges to be deployed with formal generalisation guarantees.

\end{enumerate}
\vspace{-0.4em}
\section{Related work}

\textbf{PAC-Bayes bounds.} Classical generalisation bounds based on measures such as VC-dimension \citep{vapnik1998statistical} and Rademacher complexity \citep{bartlett2002rademacher} are typically vacuous for modern over-parameterised models \citep{catoni2007pac, dziugaite2017computing}. PAC-Bayes theory instead provides data-dependent guarantees by comparing a learned posterior to a fixed prior over hypotheses \citep{mcallester1999pac, alquier2024user}, and has yielded non-vacuous bounds for deep neural networks \citep{dziugaite2017computing}. Recent work extends this to LLMs through compression-based analyses linking generalisation to empirical fit and compressibility \citep{lotfi2022pac, lotfi2024non, lotfi2024unlocking}. These works certify a model's own next-token risk and consider compression through SubLoRA \citep{lotfi2024non}, which combines LoRA with random subspace training, as well as structured Kronecker and Monarch parameterisations and post-training quantisation \citep{lotfi2024unlocking}. In contrast, we certify a judge's risk over the outputs of a separate fixed generator using tailored PEFT adaptations of pre-trained safety backbones, staying close to standard safety fine-tuning pipelines. Complementary approaches obtain distribution-free guarantees by calibration on held-out data, including conformal prediction \citep{shafer2008tutorial} and distribution-free risk control \citep{angelopoulos2025learn}. These certify a fixed predictor but depend only on held-out error and sample size, so cannot capture how certificate tightness varies with adaptation structure (\cref{app:validation-comparison}).

\textbf{Parameter-efficient fine-tuning.} Parameter-efficient fine-tuning (PEFT) methods adapt LLMs through low-dimensional parameter updates. LoRA \citep{hu2022lora} learns low-rank weight updates, while LoRA-XS \citep{balazy2024lora} learns a small latent update in the SVD-derived space of the frozen weights. LoRA-T (TinyLoRA) \citep{morris2026learning} further reduces this update using fixed random projections and a small number of learned scaling parameters, following VeRA \citep{kopiczko2024vera} in freezing random projections. These increasingly compact adaptations provide a natural setting for studying how adaptation simplicity affects predictive performance and certificate tightness.

\textbf{LLMs as safety oversight models.} Scalable oversight uses automated evaluators to supervise large models \citep{bowman2022measuring}, while the LLM-as-a-judge paradigm \citep{zheng2023judging} uses specialised models to evaluate their outputs. Models such as Llama Guard \citep{inan2023llama, chi2024llama} and Qwen3Guard \citep{zhao2025qwen3guard} act as system-level safety filters, classifying prompts and responses according to predefined risk taxonomies. Such models do not typically come with formal guarantees on performance, nor an understanding of which properties support their generalisation, leaving a gap between their practical use and reliable deployment.
\vspace{-0.4em}
\section{Preliminaries}

\subsection{Quantisation, compression, and description length}
\label{sec:quant-compression}

Quantisation maps real-valued parameters to a finite discrete representation \citep{gersho2012vector}. Given $\theta \in \mathbb{R}^p$ and bit-width $b \in \mathbb{N}$, a quantisation scheme is defined via a codebook $\mathcal{C}_b$ with $|\mathcal{C}_b| \leq 2^b$, a space of reconstruction parameters $\mathcal{S}$, and a dequantisation rule $d_b : \mathcal{C}_b^p \times \mathcal{S} \to \mathbb{R}^p$. The quantised representation is $(z,s) \in \mathcal{C}_b^p \times \mathcal{S}$, where $z$ contains the quantised values and $s$ the reconstruction parameters. The reconstructed parameters are $\hat{\theta} = d_b(z,s)$. For example, under linear scaling, $\hat{\theta} = sz$ for a scale $s > 0$. Data-driven schemes such as GPTQ \citep{frantar2022gptq} and QuIP \citep{tseng2024quip} choose $z$ to limit changes in model outputs, while for very small parameterisations nearest-integer rounding suffices. The resulting discrete representation can be further compressed losslessly without changing the reconstructed parameters. We encode it using a lossless compressor such as Brotli \citep{google_brotli_v020}, producing a binary string $c(h) \in \{0,1\}^\ast$ \citep{cover1999elements}. We define the description length of a hypothesis $h$ as $|h| \coloneqq |c(h)|$, i.e. the length of the encoded binary string. This reflects the minimum description length (MDL) principle, an application of Occam's razor that favours simple hypotheses that fit the data \citep{grunwald2007minimum}. In our PAC-Bayes certificates, shorter descriptions contribute a smaller complexity term (\Cref{sec:pac-bayes-compression}).

\subsection{PAC-Bayes bounds}
\label{sec:pac-bayes}

Let $\mathbb{P}_{\mathcal{D}} \in \mathcal{P}(\mathcal{X} \times \mathcal{Z})$ denote a data distribution over example--label pairs and let $\mathcal{H}$ denote a hypothesis class of predictors $h : \mathcal{X} \to \widehat{\mathcal{Z}}$. Given a dataset $S = \{(x_i, z_i)\}_{i=1}^n \sim \mathbb{P}_{\mathcal{D}}^n$ of $n$ i.i.d.\ samples and a loss $\ell : \widehat{\mathcal{Z}} \times \mathcal{Z} \to [0,1]$, define the population and empirical risks of $h \in \mathcal{H}$ as
\begin{equation*}
\mathcal{R}_{\mathcal{D}}(h) = \mathbb{E}_{(x,z)\sim \mathbb{P}_{\mathcal{D}}}\big[\ell(h(x), z)\big],
\qquad
\widehat{\mathcal{R}}_S(h) = \frac{1}{n}\sum_{i=1}^n \ell(h(x_i), z_i).
\end{equation*}
For $\mathbb{Q}, \mathbb{P} \in \mathcal{P}(\mathcal{H})$ with $\mathbb{Q} \ll \mathbb{P}$, the Kullback--Leibler (KL) divergence is
$
\mathrm{KL}(\mathbb{Q}\|\mathbb{P}) = \int_{\mathcal{H}} \log \frac{d\mathbb{Q}}{d\mathbb{P}}(h)\, d\mathbb{Q}(h),$
where $d\mathbb{Q}/d\mathbb{P}: \mathcal{H} \to \mathbb{R}_{\geq0}$ is the Radon--Nikodym derivative of $\mathbb{Q}$ with respect to $\mathbb{P}$ \citep{billingsley1995probability}. For a posterior $\mathbb{Q}$, the Gibbs risks are
\begin{equation*}
\mathcal{R}_{\mathcal{D}}(\mathbb{Q}) = \mathbb{E}_{h \sim \mathbb{Q}}\big[\mathcal{R}_{\mathcal{D}}(h)\big],
\qquad
\widehat{\mathcal{R}}_S(\mathbb{Q}) = \mathbb{E}_{h \sim \mathbb{Q}}\big[\widehat{\mathcal{R}}_S(h)\big].
\end{equation*}
Among the many bounds in the PAC-Bayes literature \citep{alquier2024user}, we adopt the following generalised Catoni-style bound \citep{rodriguez2024more}, not explored in prior compression-based analyses of LLMs, which yields the tightest certificates in our setting (\Cref{sec:alt_bounds}).

\begin{theorem}[{\citealp[Theorem 6]{rodriguez2024more}}] \label{thm:catoni}
Let $\ell: \widehat{\mathcal{Z}} \times \mathcal{Z} \to [0,1]$ and let $\mathbb{P}$ be independent of $S$. For any $\beta \in (0,1)$, with probability at least $1-\beta$ over $S \sim \mathbb{P}_{\mathcal{D}}^n$, simultaneously for all $\mathbb{Q} \in \mathcal{P}(\mathcal{H})$ with $\mathbb{Q} \ll \mathbb{P}$,
\begin{equation*}
\mathcal{R}_{\mathcal{D}}(\mathbb{Q}) \leq \inf_{\lambda>0} \left\{ \frac{1}{1 - e^{-\frac{\lambda}{n}}} \left[ 1 - \exp\left(-\frac{\lambda \widehat{\mathcal{R}}_S(\mathbb{Q}) + \mathrm{KL}(\mathbb{Q}\|\mathbb{P}) + \log \frac{2+\sqrt{2n}}{\beta}}{n} \right) \right] \right\}.
\end{equation*}
\end{theorem}

\textbf{On optimising $\lambda$.~~} The infimum is \emph{inside} \cref{thm:catoni}, which holds simultaneously for all $\lambda>0$ on a single $1-\beta$ event. Minimising over $\lambda$ therefore evaluates the bound without a union-bound penalty or held-out data, unlike the fixed-$\lambda$ Catoni bound used in prior compression-based LLM analyses \citep{catoni2007pac,lotfi2022pac}. It is also equivalent to the Seeger--Langford bound, evaluating the same value by minimisation over $\lambda$ rather than binary-KL inversion (\cref{sec:alt_bounds}).

\subsection{Compression-based bounds}
\label{sec:pac-bayes-compression}

Following prior work on compression-based PAC-Bayes bounds for LLMs \citep{lotfi2022pac, lotfi2024unlocking}, we adopt a Solomonoff prior \citep{solomonoff1964formal} of the form
\begin{equation*}
\mathbb{P}_{\mathrm{Sol}}(h) = 2^{-K_p(h\mid M)}/{Z},
\qquad
Z = \sum_{h \in \mathcal{H}} 2^{-K_p(h\mid M)} \leq 1,
\end{equation*}
where $K_p(h\mid M)$ is the prefix-free Kolmogorov complexity of $h \in \mathcal{H}$ given decoder $M$ \citep{li2008introduction} and the bound on $Z$ is Kraft's inequality \citep{kraft1949device}. Here $M$ is the fixed, data-independent reconstruction pipeline (frozen base model, seeded adaptor parameterisation, dequantisation rule $d_b$, lossless decompressor, see \cref{sec:quantisation_details}), so $h$ is identified with the compressed adaptor code it decodes, of length $|h| = |c(h)|$, and $\mathbb{P}_{\mathrm{Sol}}$ is a valid PAC-Bayes prior, being independent of the sample $S$ over which the bound is computed.

Although $K_p$ is uncomputable, the length of any valid prefix-free encoding of $h$ under $M$ upper-bounds it, yielding a computable certificate (\cref{cor:solomonoff_guard_base}). This is the standard minimum description length construction \citep{grunwald2007minimum}, in which the encoding makes the uncomputable term computable. We use the Elias $\delta$ code, which gives a tighter description-length bound than the Elias $\gamma$ code used in prior work \citep{lotfi2024non}. Quantisation and lossless compression then determine the description length used in the certificate. The specific quantisation schemes are described in \Cref{sec:quantisation_details}.

\vspace{-0.4em}
\section{PAC-Bayes certification of large language model safety oversight}
\label{sec:judge-guarantees}

Let $\mathcal{V}$ be a finite vocabulary, with prompt space $\mathcal{X} \subseteq \mathcal{V}^*$ and response space $\mathcal{Y} \subseteq \mathcal{V}^*$. Let $\mathbb{P}_{\mathcal{D}} \in \mathcal{P}(\mathcal{X})$ be the marginal distribution over prompts, and let a fixed generator $G$ define a Markov kernel $\mathbb{P}_G(\cdot \mid x) \in \mathcal{P}(\mathcal{Y})$. This induces a joint distribution $\mathbb{P}_{\mathcal{D}, G} \in \mathcal{P}(\mathcal{X} \times \mathcal{Y})$ given by
\begin{equation*}
\mathbb{P}_{\mathcal{D},G}(B) = \int_{\mathcal{X}} \mathbb{P}_G(B_x \mid x)\, d\mathbb{P}_{\mathcal{D}}(x),
\qquad
B_x = \{y \in \mathcal{Y} : (x,y) \in B\},
\end{equation*}
and let $(x_i,y_i)_{i=1}^{n} \sim (\mathbb{P}_{\mathcal{D},G})^n$ be $n$ i.i.d.\ prompt--response pairs.

Here \textit{fixed} refers to the fact that the generator weights, decoding procedure, system prompt, and prompt distribution remain unchanged between certification and deployment. $G$ may also combine multiple models. The i.i.d.\ assumption requires that the certification sample and future deployment samples come from the same distribution, which may include adversarial and multi-turn prompts, as in Egida \citep{garcia2025efficient}.

Let $z^*: \mathcal{X} \times \mathcal{Y} \to \mathcal{Z} = \{0,1\}$ denote a deterministic ground-truth evaluation function, indicating whether a prompt--response pair satisfies the evaluation criterion, which the judge $h : \mathcal{X} \times \mathcal{Y} \to \widehat{\mathcal{Z}}$ in $\mathcal{H}_J$ seeks to emulate, with $\widehat{\mathcal{Z}} = \{0,1\}$ under $0$--$1$ loss and $\widehat{\mathcal{Z}} = [0,1]$ under Brier. Labelling gives $S = \{(x_i,y_i,z_i)\}_{i=1}^{n}$, $z_i = z^*(x_i,y_i)$. We define the judge's risk and empirical risk with respect to $\mathbb{P}_{\mathcal{D}, G}$ as
\begin{equation*}
\mathcal{R}_{G}(h) \coloneqq \mathbb{E}_{(x,y)\sim \mathbb{P}_{\mathcal{D}, G}}\big[\ell(h(x,y), z^*(x,y))\big],
\qquad
\widehat{\mathcal{R}}_{S}(h) \coloneqq \frac{1}{n} \sum_{i=1}^n \ell(h(x_i,y_i), z^*(x_i,y_i)).
\end{equation*}
In line with prior compression-based analyses of language models \citep{lotfi2024non, lotfi2024unlocking}, we restrict attention to deterministic posteriors $\mathbb{Q}_h = \delta_h$, where $\delta_h \in \mathcal{P}(\mathcal{H})$ is the Dirac measure at $h$. Since safety systems typically deploy a single fixed judge to monitor a given generator, this lets the certificate apply directly to the deployed judge without requiring Gibbs sampling from a stochastic posterior (\Cref{app:detpost}), making it simpler to apply and consistent with common practitioner usage.

\begin{proposition}\label{prop:kl-ub}
Assume $|\mathcal{H}_J| < \infty$, let $M$ be the fixed decoder above, and let $|h|\ge 1$. For $\mathbb{Q}_h = \delta_h$ and the Solomonoff prior $\mathbb{P}_{\mathrm{Sol}}$ defined above, $\mathrm{KL}(\mathbb{Q}_h \,\|\, \mathbb{P}_{\mathrm{Sol}}) \le \mathrm{KL}^{\mathrm{UB}}(\mathbb{Q}_h \,\|\, \mathbb{P}_{\mathrm{Sol}})$, where
\begin{equation*}
\mathrm{KL}^{\mathrm{UB}}(\mathbb{Q}_h \,\|\, \mathbb{P}_{\mathrm{Sol}}) = \Bigl(|h| + \lfloor\log_2|h|\rfloor + 2\bigl\lfloor\log_2(\lfloor\log_2|h|\rfloor + 1)\bigr\rfloor + 1\Bigr)\log 2.
\end{equation*}
\end{proposition}

\begin{proof}
See \Cref{sec:proof-of-deterministic_KL}.
\end{proof}

For our deterministic-posterior construction, we take the hypothesis space to be finite, as ensured by quantisation, so that the KL divergence is well-defined for point-mass posteriors. Substituting \Cref{prop:kl-ub} into \Cref{thm:catoni} yields a computable certificate.

\begin{corollary}[Oversight Model Certificate] \label{cor:solomonoff_guard_base}
Let $S \sim (\mathbb{P}_{\mathcal{D}, G})^n$. For any $\beta \in (0,1)$, with probability at least $1-\beta$ over $S$, simultaneously for all $h \in \mathcal{H}_J$,
\begin{equation*}
\mathcal{R}_{G}(h) \leq \inf_{\lambda>0} \left\{ \frac{1}{1 - e^{-\frac{\lambda}{n}}} \left[ 1 - \exp\left(-\frac{\lambda \widehat{\mathcal{R}}_S(h) + \mathrm{KL}^{\mathrm{UB}}(\mathbb{Q}_h \| \mathbb{P}_{\mathrm{Sol}}) + \log \frac{2+\sqrt{2n}}{\beta}}{n} \right) \right] \right\}.
\end{equation*}
\end{corollary}

\begin{proof}
\Cref{thm:catoni} holds simultaneously for all $\mathbb{Q} \ll \mathbb{P}_{\mathrm{Sol}}$ on a single $1-\beta$ event. Since $\mathbb{P}_{\mathrm{Sol}}(h)>0$ for every $h\in\mathcal{H}_J$, each $\mathbb{Q}_h=\delta_h$ satisfies this condition. The result follows from \Cref{prop:kl-ub}.
\end{proof}

The certificate thus bounds the risk of the deployed judge under $\mathbb{P}_{\mathcal{D},G}$. Its tightness reflects a trade-off between the empirical performance and simplicity of the learned judge, both of which are determined by the learned hypothesis and therefore depend on the certification sample. Following prior work, we fix $\beta = 0.05$ throughout.

\textbf{Assumptions and scope.}
\Cref{cor:solomonoff_guard_base} requires a sample-independent prior and decoder $M$, including the backbone, an i.i.d.\ certification sample $S \sim \mathbb{P}_{\mathcal{D},G}$, and a loss $\ell \in [0,1]$. No assumptions are made on the learning algorithm itself. The bound holds uniformly over $\mathcal{H}_J$, allowing selection after observing $S$, provided the description includes any configuration information needed for reconstruction. Recording which of $K$ configurations was chosen takes $\lceil\log_2 K\rceil$ bits (\Cref{app:certconstruct}). More generally, the certificate applies to any finite hypothesis reconstructible from an explicit code, not only PEFT. Deployment changes that alter the certified distribution require recomputation (\Cref{sec:discussion,app:scope}).

\subsection{LoRA-GT: global-scale quantised adaptations}
\label{sec:global-scale}

The Solomonoff prior assigns greater prior probability to hypotheses with shorter descriptions, formalising an Occam's razor-style preference for simpler hypotheses. This connects directly to our motivating question of whether simple adaptations of capable guardrails can retain strong predictive performance while admitting tight generalisation guarantees without reserving scarce safety data for validation. PEFT provides a natural setting for studying this question by expressing adaptations as low-dimensional updates to a frozen base model. Their complexity can then be measured through description length after quantisation and compression.

Our hypothesis class is accordingly the set of PEFT adaptations of a frozen oversight model, namely LoRA \citep{hu2022lora}, LoRA-XS \citep{balazy2024lora}, and LoRA-T \citep{morris2026learning}, of increasing parameter efficiency and defined in \Cref{sec:peft-variants}. Since \Cref{cor:solomonoff_guard_base}'s complexity term is governed by adaptor description length, increasingly compact adaptations offer the potential for tighter certification if they also retain strong predictive performance. LoRA-T is the most compact of these parameterisations, with its weight-tied variant the most compact of all. We denote the non-tied variant LoRA-TN and the weight-tied variant LoRA-TT, both due to \citet{morris2026learning}, and introduce LoRA-GT, the tied variant under global-scale quantisation, novel to this work.

In the tied variant, a single vector is shared across all adapted modules within each transformer layer. For a base weight $W_0 \in \mathbb{R}^{d \times k}$, the adapted weights of a particular weight matrix take the form
$W = W_0 + U \Sigma \left( \sum_{i=1}^u v_i P_i \right) V^\top,$
where $U, \Sigma, V$ and $\{P_i\}_{i=1}^u$ are fixed and $\mathbf{v} \in \mathbb{R}^u$ is trainable, with $\mathbf{v}^{(\ell)} \in \mathbb{R}^u$ shared within each layer $\ell \in \{1,\dots,L\}$. See \Cref{sec:peft-variants} for further details.

\textbf{Scale overhead.~~} Under $b$-bit quantisation of LoRA-TT, the per-layer scales stored in bfloat16 cost $16L$ bits, dominating $|h|$, and hence the KL penalty of \Cref{prop:kl-ub}, for small $u$ and $b$. At $u{=}1$, $b{=}2$ on Llama-Guard-3-8B ($L{=}32$), the weights require only $64$ bits while the scales require $512$. Sharing a single global scale across layers removes $(L-1)\times16$ bits of this metadata, taking the total from $576$ to $80$ bits, an $86\%$ reduction, and leaves the complexity term set more by the weights themselves than by reconstruction overhead.

\textbf{Scale selection.~~}With only $u$ trainable parameters per layer, nearest-integer rounding suffices because the representation is small enough that more elaborate second-order schemes such as GPTQ offer little benefit. We use the symmetric codebook $\mathcal{C}_b = \{-q_{\max}, \dots, q_{\max}\}$, where $q_{\max} = 2^{b-1}-1$, and a single global scale $s>0$. Writing $\mathbf{v} \in \mathbb{R}^{Lu}$ for the concatenation of all layer vectors, the quantised values are $z_i = \mathrm{clip}\bigl(\operatorname{round}(v_i/s),-q_{\max},q_{\max}\bigr)$ with reconstruction $\hat{\mathbf{v}}=sz$, where $\mathrm{clip}(x,a,b)=\min\{b,\max\{a,x\}\}$ and $\operatorname{round}$ rounds to the nearest integer. The natural clipping-free scale is $s^*=\|\mathbf{v}\|_\infty/(q_{\max}+\tfrac12)$, the smallest scale that avoids clipping and, among clipping-free choices, minimises the worst-case rounding error bound.\vspace*{-3pt}

\begin{proposition}[Minimal Clipping-Free Scale]
\label{prop:scale}
Fix $b \in \mathbb{N}$ and let $\mathbf{v} \in \mathbb{R}^{Lu}$ be nonzero. Under nearest-integer rounding with ties resolved toward zero, $s^*$ is the smallest scale for which $|\operatorname{round}(v_i/s^*)| \le q_{\max}$ for all $i$. Moreover, for any clipping-free scale $s$, $\|\mathbf{v}-\hat{\mathbf{v}}(s)\|_\infty \le s/2$, with equality whenever some $v_i/s$ is a half-integer. Hence $s^*$ minimises the worst-case rounding error bound among clipping-free scales.
\end{proposition}
\vspace*{-1.2em}
\begin{proof}
See \Cref{sec:proof-of-global-scale}.
\end{proof}
\vspace*{-5pt}

At low bit-widths, the smallest clipping-free scale need not minimise the overall reconstruction error. For example, when $b=2$, the codebook is $\mathcal{C}_2=\{-1,0,1\}$ and $s^*=\|\mathbf{v}\|_\infty/(1+\frac{1}{2})$. Thus, any coordinate with $|v_i|<s^*/2=\|\mathbf{v}\|_\infty/3$ rounds to zero. If many coordinates fall below this threshold, their errors can accumulate in the overall $\ell_2$ reconstruction error. A smaller scale narrows both the spacing $s$ between reconstruction levels and their range $[-sq_{\max},sq_{\max}]$. This lowers the threshold $s/2$ for rounding to zero, preserving more small coordinates as nonzero values at the cost of clipping the largest coordinates. We therefore start from $s^*$ and, for $b\leq 4$, search over smaller representable scales, selecting the one that minimises $\|\mathbf{v}-d_b(z(s),s)\|_2$. Because $s$ is stored in bfloat16, only finitely many scales are representable. The search rule is fixed in advance, and the selected scale is included in the encoded hypothesis. For $b\geq 8$, we use $s^*$ directly, since the finer codebook makes the clipping-free scale sufficiently accurate (\Cref{app:scale-search-bits}). \Cref{sec:distortion} also reports results using $s^*$ as a control.

\vspace{-0.4em}
\section{Empirical evaluation}
\label{sec:experiments}

\begin{figure}[t]
\vspace{-3pt}
    \centering

    \includegraphics[width=1.0\linewidth]{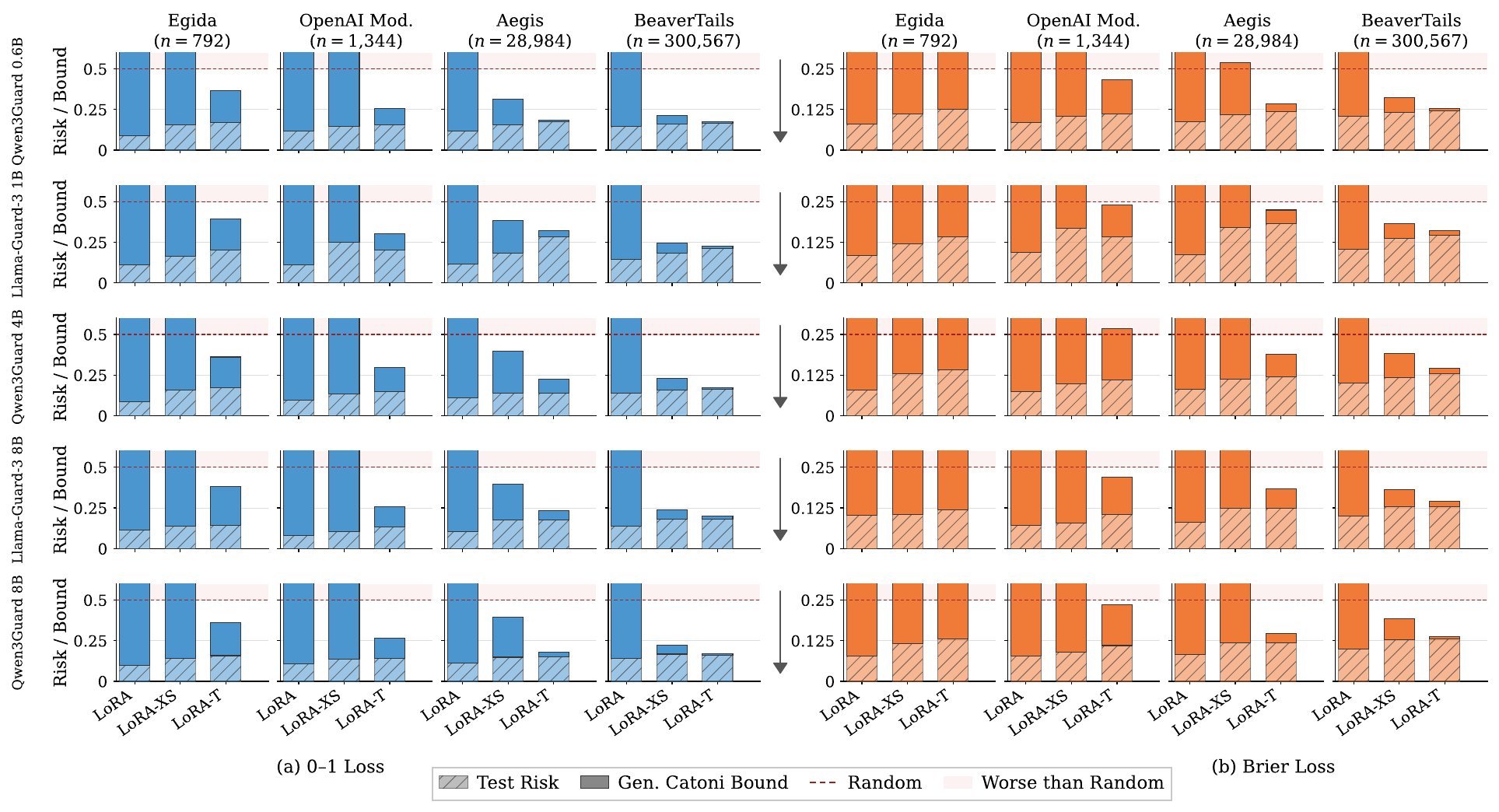}

\vspace{-8pt}
\caption{
\textbf{Tight non-vacuous guarantees for safety oversight models.}
Certified bounds and test risk for $0$--$1$ and Brier losses across models, datasets, and adaptor families, showing the tightest certificate within each family after accounting for the finite selection cost (\Cref{app:certconstruct}). LoRA, whose bounds are vacuous, is shown at its lowest test risk. Highly compressed adaptors achieve tight non-vacuous guarantees at a small cost in test risk.
}\label{fig:main-results}
\vspace{-8pt}
\end{figure}

\textbf{Models and datasets.~~} Our main results cover five safety oversight models from two guardrail families, Llama Guard 3 at 1B and 8B \citep{grattafiori2024llama} and Qwen3Guard at 0.6B, 4B and 8B \citep{zhao2025qwen3guard}. All five appear in \Cref{fig:main-results} and the complete per-model results tables (\Cref{app:detailed-results}). We use four benchmarks, Egida \citep{garcia2025efficient} ($n{=}792$), OpenAI Content Moderation \citep{markov2023holistic} ($n{=}1{,}344$), Aegis \citep{ghosh-etal-2025-aegis2} ($n{=}28{,}984$), and BeaverTails \citep{ji2023beavertails} ($n{=}300{,}567$). Each instantiates a particular $\mathbb{P}_{\mathcal{D},G}$, with the generator underlying each recorded in \Cref{sec:oversight-connection}. Our judges evaluate binary safety classification generatively, emitting tokens from the model vocabulary $\mathcal{V}$, and the predictive distributions over safety categories used for Brier scores are described in \Cref{sec:pre-distributions-over-safety-categories}.

\textbf{Fine-tuning setup.~~} We fine-tune with LoRA at ranks $r \in \{1,2,4\}$, LoRA-XS at $r \in \{2,4\}$, and the three LoRA-T variants of \Cref{sec:global-scale}, LoRA-TN and LoRA-TT at $(r,u) \in \{(2,1),\allowbreak (4,1),\allowbreak (4,2),\allowbreak (4,4)\}$ and LoRA-GT at $r = 4$ and $u \in \{1,2,4\}$. Models are trained in a standard way, simply autoregressively via supervised fine-tuning using the standard log-loss objective, adapting all linear layers. Further details are given in \Cref{sec:peft_details}.

\textbf{Quantisation and compression.~~} Following \Cref{sec:quant-compression}, we quantise adaptor parameters post-training at $2$ and $4$ bits, extending to $\{2,3,4,8,16\}$ for the distortion analysis of \Cref{sec:distortion}. Selecting the tightest certificate among $K$ bit-widths adds $\lceil\log_2 K\rceil$ selection bits, giving $1$ bit for two choices and $3$ for five (\Cref{app:certconstruct}). All variants use Hadamard incoherence processing, followed by GPTQ via \texttt{llm-compressor} \citep{llmcompressor2024} for LoRA and LoRA-XS and nearest-integer rounding for LoRA-TN, LoRA-TT and LoRA-GT, with the LoRA-GT scale chosen as in \Cref{sec:global-scale} via a scale search. Only the adaptor and its reconstruction metadata enter $|h|$ (\Cref{sec:quantisation_details}), and we retain the shorter of the raw and Brotli encodings \citep{alakuijala2015brotli_comparison}, charging one flag bit to identify the choice (\Cref{app:certconstruct,tab:compressors}). \Cref{sec:pipeline} summarises the full training and certification pipeline.

\subsection{Non-vacuous and tight guarantees for safety oversight models}
\label{sec:main-results}

\Cref{fig:main-results} shows non-vacuous guarantees under both $0$--$1$ and Brier loss across models, datasets, and adaptor families. Notably, this extends to Egida and OpenAI Moderation with only $792$ and $1{,}344$ certification samples, reaching substantially smaller sample sizes than prior compression-based LLM certificates and demonstrating certification in the common limited-data regime. The Brier results further provide guarantees on the quality of the predictive probabilities used for thresholded safety decisions, which have not been considered in prior compression-based LLM certification.

\begin{figure}[t]
    \centering

    \includegraphics[width=1.0\linewidth]{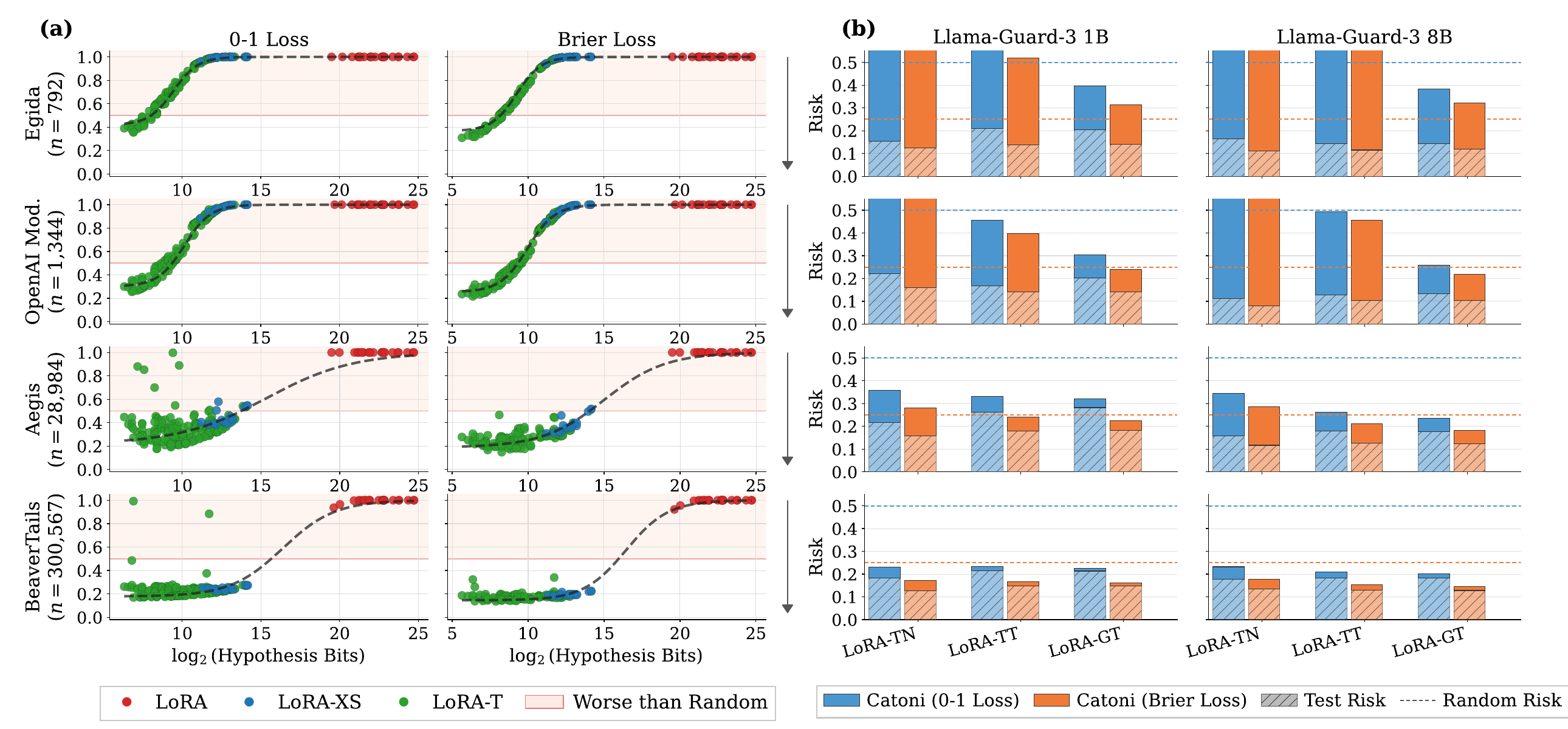}
\vspace{-0.3em}
\caption{
\textbf{Compression strongly influences certifiability in safety oversight models.}
(a) Bound vs.~description length across models and datasets, showing that smaller, more compressed adaptations yield tighter PAC-Bayes guarantees. (b) LoRA-T variants, where tying and global-scale quantisation reduce description length and tighten bounds with only modest changes in test risk.
}
    \label{fig:compression}
\vspace{-0.5em}
\end{figure}

The strongest certificates consistently come from the simplest adaptations, as also illustrated within the LoRA-T family in \Cref{fig:compression}(b). Across the complete results, LoRA-GT is tightest in all $40$ model--dataset--loss cases (\Cref{app:detailed-results,app:tinylora-variants}), while standard LoRA is vacuous in every case. This continues even on BeaverTails with $n\approx300$k, showing that large certification samples alone do not ensure tight guarantees. The complexity term depends on $|h|/n$ rather than on $n$ alone, and LoRA descriptions there take $0.8$M to $28$M bits.

\vspace*{-1pt}
\takeaway{Compact safety adaptations yield non-vacuous, often tight guarantees for classification and Brier risk, even with only $792$ certification samples.}
\vspace*{-12pt}

\clearpage

\subsection{Adaptor compression yields the tightest PAC-Bayes guarantees}
\label{sec:compression-results}

\Cref{fig:compression}(a) shows that certificate tightness closely tracks adaptor description length across models and datasets. The description-length ordering, LoRA, LoRA-XS, LoRA-TN, LoRA-TT, then LoRA-GT, matches the certificate ordering in $39$ of $40$ model--dataset--loss cases (\Cref{sec:desc-lengths}), despite the large variation in backbone scale. Within the LoRA-T family, \Cref{fig:compression}(b) shows that tying and LoRA-GT progressively shorten descriptions and tighten certificates. Adaptation complexity is therefore a stronger determinant of certificate tightness than backbone scale.

Greater compression can weaken predictive performance, although the effect is modest in many settings and varies across models and benchmarks (\Cref{app:detailed-results}). On BeaverTails, for example, quantised LoRA-GT has $2.1$ to $6.7$ points higher $0$--$1$ test risk than LoRA, while its certificate falls from $1.000$ to between $0.169$ and $0.226$ (\Cref{tab:pipeline-beavertails}). Most of the gap between LoRA and LoRA-GT is already present before quantisation, indicating that it comes from the more restricted parameterisation rather than quantisation. The adaptors therefore span a trade-off between performance and certifiability, with LoRA-GT giving substantially tighter guarantees at a generally slight cost to predictive performance. \Cref{sec:distortion} examines how far compression can go before distortion outweighs the gain from shorter descriptions.

\takeaway{Adaptor description length tracks certificate tightness more closely than backbone scale. LoRA-GT gives the tightest certificates, generally at a small cost in predictive performance.}

\subsection{How far compression can go}
\label{sec:distortion}

Let $h,\tilde h\in\mathcal{H}_J$ denote a fine-tuned judge and its quantised version, respectively. We define \emph{functional distortion} as the change in risk caused by quantisation:
\begin{equation*}
\Delta_G(h,\tilde h)
= \mathbb{E}_{(x,y)\sim\mathbb{P}_{\mathcal{D},G}}\!\left[
\ell(\tilde h(x,y),z^*(x,y))-\ell(h(x,y),z^*(x,y))
\right].
\end{equation*}
Replacing the expectation by the average over the $n$ examples in the certification sample $S$ gives the \emph{empirical functional distortion}, $\delta_n(h,\tilde h)=\widehat{\mathcal{R}}_S(\tilde h)-\widehat{\mathcal{R}}_S(h)$. Positive values mean that quantisation increases empirical risk. Compression tightens certificates when the complexity savings from shorter adaptor descriptions outweigh any increase in empirical risk stemming from increased functional distortion introduced through quantisation (\Cref{prop:distortion-complexity-decomposition}). Further details and results relating to functional distortion are given in \Cref{app:functional-distortion}.

\begin{figure}[t]
    \centering
    \includegraphics[width=1.0\linewidth]{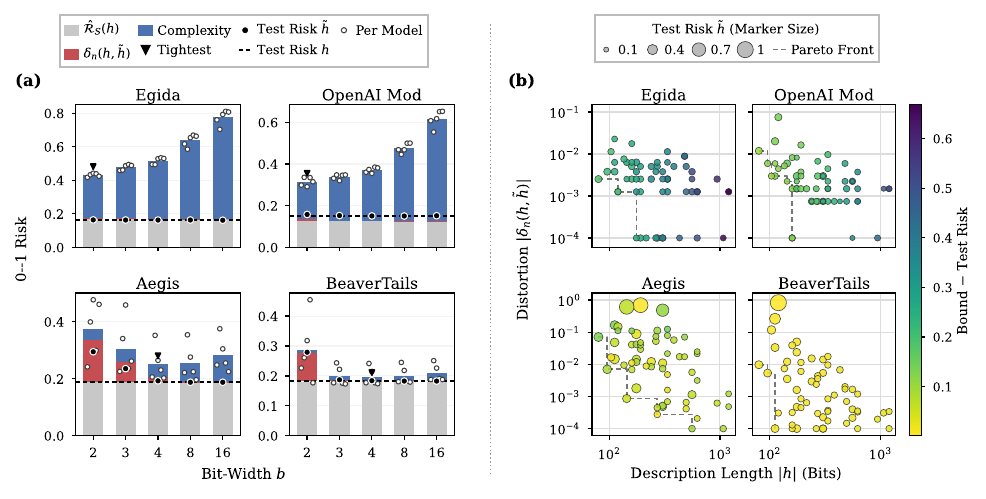}
\vspace{-0.5em}
\caption{\textbf{Compression and distortion under $0$--$1$ loss.} LoRA-GT across five models and $u\in\{1,2,4\}$. (a) Mean Catoni certificates split into pre-quantisation empirical risk, signed distortion and complexity. (b) Absolute distortion against description length at $b\in\{2,3,4,8\}$. Colour shows certificate minus test risk and marker size shows test risk. Dashed lines trace the Pareto front.}
    \label{fig:decomposition}
\vspace{-1em}
\end{figure}

The trade-off between complexity savings and functional distortion depends on the certification sample size. In \Cref{fig:decomposition}(a), mean LoRA-GT certificates under $0$--$1$ loss favour $2$ bits on Egida and OpenAI Moderation, but $4$ bits on Aegis and BeaverTails. The complexity penalty depends on $|\tilde h|/n$, the ratio of quantised adaptor description length to certification sample size. On these larger benchmarks, further compression therefore saves less, and the added functional distortion at $2$ bits loosens the certificate. In \Cref{fig:decomposition}(b), certificate tightness tracks description length on the smaller benchmarks. On the larger benchmarks, the smaller complexity penalty gives more uniform tightness across description lengths, even as test risk varies. The generalisation bound therefore depends more on the fine-tuned judge's empirical risk and the functional distortion caused by quantisation than on the complexity penalty.

\takeaway{Aggressive quantisation helps most when certification data are scarce. With more data, quantisation error can outweigh the smaller complexity savings and loosen certificates.}

\vspace{-0.4em}
\section{Discussion}
\label{sec:discussion}

Our results show that simple PEFT updates can adapt capable safety judges to new safety data while retaining strong predictive performance and admitting tight generalisation guarantees. Across models and datasets, certificate tightness tracks adaptation complexity more closely than the scale of the underlying judge. This suggests that simple adaptations can suffice to specialise existing guardrails to new safety taxonomies or guidelines, making adaptation complexity a useful consideration alongside predictive performance.

These guarantees fit within a standard fine-tuning pipeline. The adaptor is trained conventionally and certified post hoc, without constraining the learning algorithm or requiring a separate validation split. The full labelled sample can therefore be used for adaptation and certification, which is particularly relevant when safety data are scarce.

Compression can further tighten certificates by shortening adaptor descriptions, but only while the resulting functional distortion remains small. At $2$ bits, the balance depends on sample size, with compression tightening certificates on smaller benchmarks but loosening them on larger ones. More generally, our framework is not specific to PEFT and applies to any finite hypothesis reconstructible from an explicit code.

\clearpage

\textbf{Limitations.}
Our guarantees apply to the generator-induced deployment distribution from which the certification data are drawn. Changes to the generator, decoding configuration, system prompt, or prompt distribution constitute distribution shift, a longstanding challenge for generalisation guarantees \citep{ben2010theory}, which remains beyond the scope of this work. We evaluate guardrail models only up to $8$B parameters, a relevant scale for low-cost oversight of larger systems. Whether compact adaptations at substantially larger scales retain low quantisation-induced functional distortion, and admit similarly tight guarantees, remains an important direction for future work.

\vspace{-0.5em}
\section{Conclusion}

We formalised PAC-Bayes certification for safety judges evaluating outputs from a fixed deployed generator. Across five guardrail models and four benchmarks, tailored PEFT yielded non-vacuous, often tight guarantees for both classification and Brier risk, including in small-data regimes and without a separate held-out validation set. Certificate tightness tracked adaptation complexity more closely than backbone scale, with LoRA-GT giving the tightest certificates in all 40 model--dataset--loss cases. Compression can tighten certificates further, provided its reduction in description length outweighs the resulting functional distortion. These results show that simple, compressible adaptations can make formal certification practical when specialising safety judges to new taxonomies with scarce labelled data.

\section*{Acknowledgements}
TGJR acknowledges support from the Foundational Research Grants program at Georgetown University's Center for Security and Emerging Technology. Resources used in preparing this research were provided, in part, by the Province of Ontario, the Government of Canada through CIFAR, and companies sponsoring the Vector Institute. This work is additionally supported by the UKRI grant: Turing AI Fellowship EP/W002981/1.

\bibliographystyle{plainnat}
\bibliography{iclr_references,preprint_references}

\clearpage
\appendix

\crefalias{section}{appsec}
\crefalias{subsection}{appsec}
\crefalias{subsubsection}{appsec}

\setcounter{tocdepth}{2}

\setcounter{equation}{0}
\renewcommand{\theequation}{\thesection.\arabic{equation}}

\section*{\LARGE Appendix}

\vspace*{10pt}
\etocsettocdepth.toc{subsection}
\etocsettocstyle{\subsection*{Table of Contents}}{}
\tableofcontents
\clearpage

\section{Theory and analysis}
\subsection{Proofs of statements used} \label{sec:proofs}

\subsubsection{Proof of \cref{prop:kl-ub}}
\label{sec:proof-of-deterministic_KL}

\begin{proof}
The prefix-free Kolmogorov complexity $K_p(h \mid M)$ is the length of the shortest program that outputs $h$ under the fixed decoder $M$, where $M$ captures the shared decoding procedure and model specification and is therefore independent of both the data and the learned adaptation defining the hypothesis. Consequently, the length of any valid prefix-free encoding of $h$ upper-bounds $K_p(h \mid M)$.

We construct such an encoding by concatenating two components, the length $|h|$, encoded using the Elias $\delta$-code~\citep{elias1975universal}, followed by the binary string $h$ itself. This forms a valid prefix-free program for $h$. The Elias $\delta$-code for an integer $x \ge 1$ has codeword length
\[
\lfloor \log_2 x \rfloor + 2\lfloor \log_2(\lfloor \log_2 x \rfloor + 1) \rfloor + 1,
\]
and hence, for $|h| \ge 1$,
\begin{equation}
K_p(h \mid M)
\leq
|h|
+ \lfloor \log_2 |h| \rfloor
+ 2\lfloor \log_2(\lfloor \log_2 |h| \rfloor + 1) \rfloor
+ 1.
\label{eq:delta-code-kolmogorov-bound}
\end{equation}
For the deterministic posterior $\mathbb{Q}_h = \delta_h$, and assuming $|\mathcal{H}_J| < \infty$, the KL divergence simplifies to
\begin{equation}
\mathrm{KL}(\mathbb{Q}_h \| \mathbb{P}_{\mathrm{Sol}})
=
\sum_{g \in \mathcal{H}_J}
\delta_h(g)
\log \frac{\delta_h(g)}{\mathbb{P}_{\mathrm{Sol}}(g)}
= -\log \mathbb{P}_{\mathrm{Sol}}(h).
\end{equation}
Using the definition of the Solomonoff prior in \Cref{sec:pac-bayes-compression},
\begin{equation}
-\log \mathbb{P}_{\mathrm{Sol}}(h)
=
-\log \left(\frac{2^{-K_p(h \mid M)}}{Z}\right)
= K_p(h \mid M)\log 2 + \log Z
\leq K_p(h \mid M)\log 2,
\end{equation}
where the inequality follows from $Z \leq 1$ by Kraft's inequality~\citep{kraft1949device}. Substituting \eqref{eq:delta-code-kolmogorov-bound} yields
\[
\mathrm{KL}(\mathbb{Q}_h \| \mathbb{P}_{\mathrm{Sol}})
\leq
\left(
|h|
+ \lfloor \log_2 |h| \rfloor
+ 2\lfloor \log_2(\lfloor \log_2 |h| \rfloor + 1) \rfloor
+ 1
\right)\log 2,
\]
which completes the proof.
\end{proof}

\noindent The Elias $\delta$ code used here yields a tighter description-length bound than the Elias $\gamma$ code of \citet{lotfi2024non} for the adaptors considered here (\Cref{tab:elias-comparison}).

\subsubsection{Proof of \cref{prop:scale}} \label{sec:proof-of-global-scale}

\begin{proof}
Write $V_{\max} = \|\mathbf{v}\|_\infty$ for the concatenated vector $\mathbf{v} \in \mathbb{R}^{Lu}$ of \Cref{sec:global-scale}, and recall $q_{\max} = 2^{b-1}-1$. A scale $s \in \mathbb{R}_{>0}$ is clipping-free precisely when
\begin{equation*}
\left|\operatorname{round}\!\left(\frac{v_i}{s}\right)\right| \leq q_{\max} \quad \text{for all } i \in [Lu],
\end{equation*}
which holds whenever $|v_i|/s \leq q_{\max} + \tfrac{1}{2}$. Under round-half-toward-zero the boundary case $|v_i|/s = q_{\max} + \tfrac{1}{2}$ does not clip, so the smallest clipping-free scale is
\begin{equation*}
s^* = \frac{V_{\max}}{q_{\max} + \tfrac{1}{2}},
\end{equation*}
matching \Cref{sec:global-scale}.

The scale $s^*$ is itself clipping-free, since $|v_i|/s^* \leq V_{\max}/s^* = q_{\max} + \tfrac{1}{2}$ for all $i$, and under round-half-toward-zero the boundary value rounds to $q_{\max}$.

For any clipping-free $s \in \mathbb{R}_{>0}$, the reconstruction error under $d_b(z,s) = s z$ is governed by rounding:
\begin{equation*}
|v_i - \hat{v}_i(s)|
= \left|v_i - s \cdot \operatorname{round}\!\left(\frac{v_i}{s}\right)\right|
= s \left| \frac{v_i}{s} - \operatorname{round}\!\left(\frac{v_i}{s}\right) \right|
\leq \frac{s}{2},
\end{equation*}
since nearest-integer rounding incurs error at most $1/2$, giving the stated inequality at $s = s^*$. The bound is attained whenever some $v_i/s$ is a half-integer, so the worst-case error over $\mathbf{v}$ at scale $s$ is exactly $s/2$. This worst-case error increases with $s$ and is therefore minimised among clipping-free scales by the smallest such scale, $s^*$. This is a worst-case guarantee and need not minimise the reconstruction error for every fixed $\mathbf{v}$.
\end{proof}

\subsection{Scope, assumptions, and validity of the certificate}
\label{app:scope}

\subsubsection{What the certificate covers}
\textbf{The certified distribution.~~} \Cref{cor:solomonoff_guard_base} is a statement about a specific joint distribution $\mathbb{P}_{\mathcal{D},G}$, determined by the generator, the decoding procedure and sampling parameters, the system prompt, and the prompt distribution. Holding these fixed between certification and deployment is what a distribution-level generalisation statement requires. The i.i.d. condition constrains how the certification and deployment samples relate to one another and places no restriction on the content of the distribution, which may contain adversarial prompts, jailbreak attempts, and multi-turn conversations. Egida, for instance, consists of jailbreak attempts against safety-aligned models, and is certified non-vacuously in \Cref{sec:main-results}. What is not covered is a change to the distribution itself. Changing the generator, the decoding configuration, or the prompt distribution induces a different $\mathbb{P}_{\mathcal{D},G}$ and requires recomputation, which for a fixed adaptor costs one pass over a labelled sample from the new distribution.

\textbf{Independence from $S$, not from $\mathbb{P}_{\mathcal{D},G}$.~~} The backbone, prior, and decoder must be fixed independently of the certification sample $S$. They need not be independent of the data-generating distribution. Knowledge already present in the pre-trained judge is therefore permitted, and is what makes low-complexity adaptation sufficient, since the certificate measures the complexity of the learned adaptation relative to a fixed backbone rather than the complexity of the resulting judge. The benchmarks we use are standard post-training evaluation sets for these models, and we are not aware of them forming part of the backbones' training data.

\textbf{The remaining conditions.~~} Two further requirements are mild in this setting. The loss must be bounded in $[0,1]$, which holds for the $0$--$1$ loss and, since predictions lie in $[0,1]$ and labels in $\{0,1\}$, for the Brier loss too. The hypothesis space must be finite, as assumed in \Cref{prop:kl-ub}, which quantisation supplies automatically by restricting parameters to finite precision. No assumption is placed on the learning algorithm or on the optimiser, since \Cref{cor:solomonoff_guard_base} holds uniformly over $\mathcal{H}_J$.

\textbf{Beyond PEFT.~~} The certificate applies to any finite hypothesis reconstructible from an explicit code, so full-parameter fine-tuning is admissible in principle, though its description length is generally too large to yield useful bounds at the sample sizes considered here. In our setting, large-scale adaptation is typically unnecessary as the pretrained model already provides the main capability, and adapting it to a new safety taxonomy usually requires only modest changes supported by limited data. Prompt-based monitors are also admissible. A prompt fixed independently of $S$ forms part of the decoder $M$ and adds no bits. A prompt selected using $S$ defines a hypothesis choice that must be encoded and charged in the complexity term.

\subsubsection{Why optimising $\lambda$ and selecting configurations introduces no leakage}
\label{app:no-leakage}
Two aspects of our procedure use the certification sample $S$: the choice of $\lambda$ and the choice of configuration.

First, we minimise the bound of \Cref{cor:solomonoff_guard_base} over $\lambda$. The infimum is part of the theorem statement. The inequality holds simultaneously for all $\lambda > 0$ on a single $1-\beta$ event, so evaluating it at the empirically best $\lambda$ is an evaluation of a bound that was already valid uniformly, not a selection among separately valid bounds. No union bound is needed and no held-out data is used. This differs from the standard Catoni bound \citep{catoni2007pac}, which is stated for a fixed, data-independent $\lambda$ and would require a union-bound penalty to optimise. The confidence parameter $\beta$ is fixed at $0.05$ throughout and never tuned.

Second, \Cref{thm:catoni} holds simultaneously for all posteriors on a single $1-\beta$ event, so \Cref{cor:solomonoff_guard_base} holds simultaneously for all $h \in \mathcal{H}_J$ with no union bound over $\mathcal{H}_J$, and therefore places no assumption on how $h$ is obtained. Selecting an adaptor family, an $(r,u)$ setting, or a bit-width using $S$ is therefore admissible, provided the resulting hypothesis is the one whose description length is charged. Recording which of $K$ configurations was chosen takes $\lceil \log_2 K \rceil$ bits (\Cref{app:certconstruct}). No validation split is introduced at any point, so the full sample is used for both fine-tuning and certification.

\subsubsection{Deterministic posteriors}
\label{app:detpost}

The object being certified here is the deployed compressed oversight model itself rather than a sampled Gibbs classifier, so deterministic posteriors $\mathbb{Q}_h = \delta_h$ are the natural choice. The certificate applies directly to the single adaptation served at inference time as is the standard practice for judge deployments.

They are also well matched to compression-based certification. Under a Dirac posterior the KL term reduces to a complexity penalty determined entirely by the description length of the deployed adaptation under the Solomonoff prior, giving an interpretable trade-off between empirical fit and compressibility, and complexity accounting in this framework is defined at the level of individual compressed hypotheses. This matches prior compression-based PAC-Bayes analyses for LLMs \citep{lotfi2024non,lotfi2024unlocking}, so the results are directly comparable and the effects of our refinements to bound choice, description-length accounting, and compressed representation are isolated.

Optimised stochastic posteriors may in principle yield tighter certificates, but approximating Gibbs risks requires repeated sampling and evaluation. At deployment this adds inference and latency overhead especially at the scale of LLMs.

\subsection{Risk functionals and predictive distributions}
\label{sec:risk-definitions}

We consider the two bounded risk functionals induced by losses $\ell: \widehat{\mathcal{Z}} \times \mathcal{Z} \to [0,1]$ used in the main body of this work, following the notation of \Cref{sec:judge-guarantees}.

\subsubsection{Bounded risk functionals}

\textbf{0--1 risk.}
For binarised correctness-based evaluation of LLM safety oversight models, we consider the standard $0$--$1$ loss, measuring safety misclassification, which induces the risk
\begin{equation*}
\mathcal{R}_{\text{0--1}}(h) = \mathbb{E}_{(x,y)\sim \mathbb{P}_{\mathcal{D},G}} \left[\mathbf{1}\big(h(x,y) \neq z^*(x,y)\big)\right],
\end{equation*}
where $z^*(x,y) \in \{0,1\}$ is the ground-truth evaluation of a prompt--response pair and $h(x,y)$ the prediction of the judge. Equivalently, this is the probability that the judge makes an error,
\begin{equation*}
\mathcal{R}_{\text{0--1}}(h) = \mathbb{P}_{(x,y)\sim \mathbb{P}_{\mathcal{D},G}}\big(h(x,y) \neq z^*(x,y)\big).
\end{equation*}

\textbf{Brier risk.}
To evaluate probabilistic predictions we use the Brier score, a strictly proper scoring rule capturing both accuracy and calibration \citep{murphy1973new}. Under this loss the judge output being scored is a probability rather than a hard label. Let $p_h(x,y) \in [0,1]$ denote the judge's predictive probability for the class $z^* = 1$, extracted from its token-level distribution as in \Cref{sec:pre-distributions-over-safety-categories}. The induced risk is then
\begin{equation*}
\mathcal{R}_{\text{Brier}}(h) = \mathbb{E}_{(x,y)\sim \mathbb{P}_{\mathcal{D},G}} \left[\big(p_h(x,y) - z^*(x,y)\big)^2\right].
\end{equation*}

Writing $c \coloneqq z^*(x,y) \in \{0,1\}$ and conditioning on the predicted probability $p_h(x,y)$ yields the classical decomposition of \citet{murphy1973new}, with all expectations taken over $(x,y)\sim \mathbb{P}_{\mathcal{D},G}$:
\begin{align*}
\mathcal{R}_{\mathrm{Brier}}(h)
= \underbrace{\mathbb{E}\!\left[\big(p_h(x,y) - \mathbb{E}[c \mid p_h(x,y)]\big)^2\right]}_{\text{Calibration}}
\;-\;
\underbrace{\mathbb{V}\!\big(\mathbb{E}[c \mid p_h(x,y)]\big)}_{\text{Resolution}}
\;+\;
\underbrace{\mathbb{V}(c)}_{\text{Uncertainty}}.
\end{align*}

The calibration term measures the discrepancy between predicted probabilities and true conditional frequencies. The resolution term, equal to the variance of the conditional label mean $\mathbb{E}[c \mid p_h(x,y)]$, captures how informative predictions are about the label, and the uncertainty term reflects the intrinsic variability of the labels.

\subsubsection{Predictive distributions over safety categories} \label{sec:pre-distributions-over-safety-categories}
The safety oversight model $h \in \mathcal{H}_J$ is generative and outputs a sequence of tokens, so to obtain a predictive distribution over output categories (e.g.\ \texttt{safe}, \texttt{unsafe}) we extract probabilities from the model's token-level conditional distribution.

Let $\mathcal{C}$ denote the finite set of output categories, and let each $c \in \mathcal{C}$ correspond to a token sequence $w^{(c)}_{1:T_c} \in \mathcal{V}^*$. For an input $(x,y) \in \mathcal{X} \times \mathcal{Y}$, the judge induces an autoregressive distribution over token sequences, giving the likelihood
\[
s_h(c \mid x,y) = \prod_{t=1}^{T_c} p_h\big(w^{(c)}_t \mid x,y, w^{(c)}_{<t}\big)
\]
for each category, which we normalise across $\mathcal{C}$ to obtain
\[
p_h(c \mid x,y) = \frac{s_h(c \mid x,y)}{\sum_{c' \in \mathcal{C}} s_h(c' \mid x,y)}.
\]
In the common case where each category corresponds to a single token, this reduces to extracting the next-token probabilities and renormalising over $\mathcal{C}$. These are the predictive probabilities to which the Brier risk of \Cref{sec:risk-definitions} is applied, with $p_h(x,y)$ there denoting $p_h(c \mid x,y)$ for the category with $z^* = 1$, and therefore the object our Brier certificates bound.

\subsection{Functional distortion under compression}
\label{app:functional-distortion}

Functional distortion quantifies how compression changes the behaviour of a learned judge. This section splits the certificate of \Cref{cor:solomonoff_guard_base} into the two channels through which compression acts, namely a perturbation of the empirical risk and a change in description length, and proves the monotonicity, sensitivity and sufficiency statements used in \Cref{sec:distortion} (\Cref{prop:distortion-complexity-decomposition,prop:certificate-sensitivity,prop:sufficient}).

Let $h\in\mathcal H_J$ denote a fine-tuned judge and let $\tilde h\in\mathcal H_J$ denote a compressed version of the same judge. For a bounded loss $\ell:\widehat{\mathcal Z}\times\mathcal Z\to[0,1]$, define the population functional distortion under the generator-induced distribution $\mathbb{P}_{\mathcal{D},G}$ as
\begin{align*}
\Delta_G(h,\tilde h)
&\coloneqq  \mathcal{R}_G(\tilde h)-\mathcal{R}_G(h) \\
&= \mathbb E_{(x,y)\sim \mathbb{P}_{\mathcal{D},G}}
\left[
\ell(\tilde h(x,y),z^*(x,y))
-
\ell(h(x,y),z^*(x,y))
\right].
\end{align*}
Given a sample $S=\{(x_i,y_i)\}_{i=1}^n\sim(\mathbb{P}_{\mathcal{D},G})^n$, define the empirical functional distortion as
\begin{align*}
\delta_n(h,\tilde h)
&\coloneqq
\frac{1}{n}\sum_{i=1}^n
\left[
\ell(\tilde h(x_i,y_i),z^*(x_i,y_i))
-
\ell(h(x_i,y_i),z^*(x_i,y_i))
\right] \\
&=
\widehat{\mathcal{R}}_S(\tilde h)-\widehat{\mathcal{R}}_S(h).
\end{align*}
The quantity $\delta_n(h,\tilde h)$ is signed, positive when compression increases empirical risk and negative when it decreases it. In our experiments $S$ is the sample used for certification, so $\delta_n(h,\tilde h)$ is computed directly as the difference in empirical risk between the compressed and fine-tuned judges.

\begin{lemma}[Empirical risk decomposition]
\label{lem:empirical-risk-decomposition-distortion}
For any $h,\tilde h\in\mathcal H_J$ and any sample $S$,
\[
\widehat{\mathcal{R}}_S(\tilde h)=\widehat{\mathcal{R}}_S(h)+\delta_n(h,\tilde h).
\]
\end{lemma}

\begin{proof}
Immediate from the definition of $\delta_n(h,\tilde h)$.
\end{proof}

We next isolate the dependence of the generalised Catoni certificate on empirical risk and complexity. For $\lambda>0$ define
\begin{align*}
\Phi_\lambda(\widehat r,\kappa;n,\beta)
&\coloneqq
\frac{1}{1-e^{-\lambda/n}}
\left[
1-\exp\left(
-\frac{\lambda \widehat r+\kappa+\log\frac{2+\sqrt{2n}}{\beta}}{n}
\right)
\right],
\\[0.4em]
\mathcal B(\widehat r,\kappa;n,\beta)
&\coloneqq
\inf_{\lambda>0}\Phi_\lambda(\widehat r,\kappa;n,\beta),
\end{align*}
so that \Cref{cor:solomonoff_guard_base} can be written as
\[
\mathcal{R}_G(h)
\le
\mathcal B
\left(
\widehat{\mathcal{R}}_S(h),
\mathrm{KL}^{\mathrm{UB}}(\mathbb{Q}_h\|\mathbb{P}_{\mathrm{Sol}});
n,\beta
\right).
\]

\begin{lemma}[Monotonicity of the certificate]
\label{lem:catoni-certificate-monotonicity}
For fixed $n\in\mathbb N$ and $\beta\in(0,1)$, the map $(\widehat r,\kappa) \mapsto \mathcal B(\widehat r,\kappa;n,\beta)$ is non-decreasing in both arguments.
\end{lemma}

\begin{proof}
Fix $\lambda>0$. Since $\lambda>0$ and $n>0$, the factor $(1-e^{-\lambda/n})^{-1}$ is positive, and
\begin{align*}
\frac{\partial \Phi_\lambda}{\partial \widehat r}
&=
\frac{\lambda}{n(1-e^{-\lambda/n})}
\exp\left(
-\frac{\lambda \widehat r+\kappa+\log\frac{2+\sqrt{2n}}{\beta}}{n}
\right)
\;\ge\; 0,
\\[0.4em]
\frac{\partial \Phi_\lambda}{\partial \kappa}
&=
\frac{1}{n(1-e^{-\lambda/n})}
\exp\left(
-\frac{\lambda \widehat r+\kappa+\log\frac{2+\sqrt{2n}}{\beta}}{n}
\right)
\;\ge\; 0.
\end{align*}
Hence $\Phi_\lambda$ is non-decreasing in both arguments, so if $(\widehat r_1,\kappa_1)\le(\widehat r_2,\kappa_2)$ coordinate-wise then $\Phi_\lambda(\widehat r_1,\kappa_1;n,\beta) \le \Phi_\lambda(\widehat r_2,\kappa_2;n,\beta)$ for all $\lambda>0$. Taking the infimum over $\lambda>0$ preserves the inequality.
\end{proof}

\begin{proposition}[Distortion--complexity decomposition]
\label{prop:distortion-complexity-decomposition}
Let $h\in\mathcal H_J$ be a fine-tuned judge and let $\tilde h\in\mathcal H_J$ be a compressed judge. Under the deterministic posterior $\mathbb{Q}_{\tilde h}=\delta_{\tilde h}$, the certificate for $\tilde h$ can be written as
\[
\mathcal{R}_G(\tilde h)
\le
\mathcal B\!\left(
\widehat{\mathcal{R}}_S(h)+\delta_n(h,\tilde h),\;
\mathrm{KL}^{\mathrm{UB}}(\mathbb{Q}_{\tilde h}\|\mathbb{P}_{\mathrm{Sol}});\,
n,\beta
\right)
\]
on the same high-probability event as \Cref{cor:solomonoff_guard_base}. Moreover, the right-hand side is non-decreasing in both $\delta_n(h,\tilde h)$ and $\mathrm{KL}^{\mathrm{UB}}(\mathbb{Q}_{\tilde h}\|\mathbb{P}_{\mathrm{Sol}})$.
\end{proposition}

\begin{proof}
Applying \Cref{cor:solomonoff_guard_base} to $\tilde h$ gives
\[
\mathcal{R}_G(\tilde h)
\le
\mathcal B\!\left(
\widehat{\mathcal{R}}_S(\tilde h),\;
\mathrm{KL}^{\mathrm{UB}}(\mathbb{Q}_{\tilde h}\|\mathbb{P}_{\mathrm{Sol}});\,
n,\beta
\right),
\]
and by \Cref{lem:empirical-risk-decomposition-distortion}, $\widehat{\mathcal{R}}_S(\tilde h) = \widehat{\mathcal{R}}_S(h)+\delta_n(h,\tilde h)$. Substituting this identity gives the stated bound, and the monotonicity claim follows from \Cref{lem:catoni-certificate-monotonicity}.
\end{proof}

\Cref{prop:distortion-complexity-decomposition} gives a formal interpretation of the compression--distortion trade-off observed in \Cref{sec:compression-results}. For a fixed fine-tuned judge $h$, compression affects the certificate through two terms, the empirical functional distortion $\delta_n(h,\tilde h)$, which perturbs the empirical-risk term, and the compressed description length $|\tilde h|$, which controls the complexity term through \Cref{prop:kl-ub}. Tighter certificates therefore favour compressed hypotheses that reduce $|\tilde h|$ while keeping $\delta_n(h,\tilde h)$ small and empirical risk low.

\begin{proposition}[Sensitivity of the PAC-Bayes certificate to compression]
\label{prop:certificate-sensitivity}
Write $\kappa_h \coloneqq \mathrm{KL}^{\mathrm{UB}}(\mathbb{Q}_h\|\mathbb{P}_{\mathrm{Sol}})$ and $\kappa_{\tilde h} \coloneqq \mathrm{KL}^{\mathrm{UB}}(\mathbb{Q}_{\tilde h}\|\mathbb{P}_{\mathrm{Sol}})$. For any $\lambda>0$,
\[
\Phi_\lambda\!\left(\widehat{\mathcal{R}}_S(h)+\delta_n(h,\tilde h),\kappa_{\tilde h};n,\beta\right)
\le
\Phi_\lambda\!\left(\widehat{\mathcal{R}}_S(h),\kappa_h;n,\beta\right)
+
\frac{\lambda\,|\delta_n(h,\tilde h)|+|\kappa_{\tilde h}-\kappa_h|}
{n\left(1-e^{-\lambda/n}\right)}.
\]
If the infimum defining $\mathcal B(\widehat{\mathcal{R}}_S(h),\kappa_h;n,\beta)$ is attained at some $\lambda_h^\star>0$, then
\[
\mathcal B\!\left(\widehat{\mathcal{R}}_S(h)+\delta_n(h,\tilde h),\kappa_{\tilde h};n,\beta\right)
\le
\mathcal B\!\left(\widehat{\mathcal{R}}_S(h),\kappa_h;n,\beta\right)
+
\frac{\lambda_h^\star\,|\delta_n(h,\tilde h)|+|\kappa_{\tilde h}-\kappa_h|}
{n\left(1-e^{-\lambda_h^\star/n}\right)}.
\]
\end{proposition}

\begin{proof}
Fix $\lambda>0$. Along the segment between the two evaluation points, the empirical-risk coordinate lies in $[0,1]$ and the complexity coordinate is nonnegative. Hence the exponential factor in the partial derivatives of $\Phi_\lambda$ is at most $1$, giving derivative bounds $\lambda/[n(1-e^{-\lambda/n})]$ and $1/[n(1-e^{-\lambda/n})]$. The mean value theorem along this segment gives the first inequality. Evaluating it at $\lambda_h^\star$ and using the definition of $\mathcal B$ gives the second.
\end{proof}

\Cref{prop:certificate-sensitivity} bounds the compressed judge's certificate by the original judge's certificate plus penalties for changes in empirical risk and complexity. If the coding scheme ensures $\kappa_{\tilde h}\le\kappa_h$, monotonicity removes the complexity penalty: only positive distortion can worsen the certificate, and
\[
\Phi_\lambda\!\left(\widehat{\mathcal{R}}_S(h)+\delta_n(h,\tilde h),\kappa_{\tilde h};n,\beta\right)
\le
\Phi_\lambda\!\left(\widehat{\mathcal{R}}_S(h),\kappa_h;n,\beta\right)
+
\frac{\lambda\max\{0,\delta_n(h,\tilde h)\}}{n(1-e^{-\lambda/n})}.
\]

\begin{proposition}[Sufficient condition for compression to tighten the certificate]
\label{prop:sufficient}
Let $h,\tilde h\in\mathcal H_J$, and write $\kappa_h \coloneqq \mathrm{KL}^{\mathrm{UB}}(\mathbb{Q}_h\|\mathbb{P}_{\mathrm{Sol}})$ and $\kappa_{\tilde h} \coloneqq \mathrm{KL}^{\mathrm{UB}}(\mathbb{Q}_{\tilde h}\|\mathbb{P}_{\mathrm{Sol}})$. Suppose the infimum defining $\mathcal B(\widehat{\mathcal{R}}_S(h),\kappa_h;n,\beta)$ is attained at $\lambda_h^\star>0$. If
\begin{equation}
\label{eq:sufficient}
\lambda_h^\star\,\delta_n(h,\tilde h)\le \kappa_h-\kappa_{\tilde h},
\end{equation}
then
\[
\mathcal B\!\left(\widehat{\mathcal{R}}_S(h)+\delta_n(h,\tilde h),\kappa_{\tilde h};n,\beta\right)
\le
\mathcal B\!\left(\widehat{\mathcal{R}}_S(h),\kappa_h;n,\beta\right).
\]
Thus, at $\lambda_h^\star$, any increase in empirical risk is offset by a sufficient decrease in complexity.
\end{proposition}

\begin{proof}
For fixed $\lambda>0$, $\Phi_\lambda(\widehat r,\kappa;n,\beta)$ is non-decreasing in $\lambda\widehat r+\kappa$. Condition \eqref{eq:sufficient} therefore implies
\[
\Phi_{\lambda_h^\star}\!\left(\widehat{\mathcal{R}}_S(h)+\delta_n(h,\tilde h),\kappa_{\tilde h};n,\beta\right)
\le
\Phi_{\lambda_h^\star}\!\left(\widehat{\mathcal{R}}_S(h),\kappa_h;n,\beta\right)
=
\mathcal B\!\left(\widehat{\mathcal{R}}_S(h),\kappa_h;n,\beta\right).
\]
Taking the infimum over $\lambda>0$ on the left proves the claim.
\end{proof}

A complexity reduction of $\lambda_h^\star\delta_n(h,\tilde h)$ nats is therefore sufficient to offset positive empirical distortion. To interpret this scale, let $K_h\coloneqq\kappa_h+\log((2+\sqrt{2n})/\beta)$ and write $\lambda=nt$. For fixed $\widehat{\mathcal{R}}_S(h)\in(0,1)$ and $K_h/n\to0$, the minimiser scales as $\lambda_h^\star=\Theta(\sqrt{nK_h})$. When $\kappa_h$ dominates the distribution-free overhead, this becomes $\Theta(\sqrt{n\kappa_h})$. Thus, as the certification sample grows, a given complexity reduction offsets less distortion, consistent with \Cref{sec:distortion}.

\subsection{Alternative PAC-Bayes risk certificates} \label{sec:alt_bounds}

While the main text uses the generalised Catoni bound, we also evaluate several alternative PAC-Bayes certificates from the literature in the setting of LLM safety oversight, both to compare against existing approaches applied to LLMs \citep{lotfi2024non, lotfi2024unlocking} and to measure how much of the tightness is due to the choice of bound alone.

Throughout, let $S \sim (\mathbb{P}_{\mathcal{D},G})^n$, let $\ell$ be bounded in $[0,1]$, let $\mathbb{P}_{\mathrm{Sol}}$ be independent of $S$, and fix $\beta \in (0,1)$. Each certificate below holds with probability at least $1-\beta$ over $S$, simultaneously for all $h \in \mathcal{H}_J$ under the deterministic posterior $\mathbb{Q}_h = \delta_h$, and is obtained by substituting \Cref{prop:kl-ub} into the corresponding bound from the literature, exactly as \Cref{cor:solomonoff_guard_base} does for the generalised Catoni bound. We write $\xi(n) = 2+\sqrt{2n}$ for the sample-size factor in \Cref{thm:catoni}. For $p,q \in [0,1]$, we define the binary KL divergence as
\begin{equation*}
\mathrm{kl}(q \| p) \coloneqq \mathrm{KL}\bigl(\operatorname{Bern}(q) \,\|\, \operatorname{Bern}(p)\bigr) = q \log \frac{q}{p} + (1-q) \log \frac{1-q}{1-p},
\end{equation*}
where $\operatorname{Bern}(p)$ denotes the Bernoulli distribution with success probability $p$, with boundary values defined by limits. We use $\mathrm{kl}$ for this binary divergence and $\mathrm{KL}$ for the divergence between probability distributions, including those over $\mathcal{H}_J$.

\begin{proposition}[Lotfi Finite Hypothesis Certificate] \label{prop:lotfi_cert}
For discrete, finite hypothesis spaces \citep{lotfi2024non},
\begin{equation*}
\mathcal{R}_G(h) \leq \widehat{\mathcal{R}}_S(h)+ \sqrt{\frac{\mathrm{KL}^{\mathrm{UB}}(\mathbb{Q}_h \| \mathbb{P}_{\mathrm{Sol}}) + \log(1/\beta)}{2n}} \,.
\end{equation*}
\end{proposition}

\begin{proposition}[McAllester Certificate] \label{prop:mcallester_cert}
The classic PAC-Bayes bound \citep{mcallester1999pac} gives
\begin{equation*}
\mathcal{R}_G(h) \leq \widehat{\mathcal{R}}_S(h)+ \sqrt{\frac{\mathrm{KL}^{\mathrm{UB}}(\mathbb{Q}_h \| \mathbb{P}_{\mathrm{Sol}}) + \log(\xi(n)/\beta)}{2n}} \,.
\end{equation*}
\end{proposition}

\begin{proposition}[Seeger--Langford Certificate] \label{prop:seeger_cert}
From the PAC-Bayes theorem for Bernoulli random variables \citep{seeger2002pac},
\begin{equation*}
\mathrm{kl}\bigl(\widehat{\mathcal{R}}_S(h) \,\big\|\, \mathcal{R}_G(h)\bigr) \leq \frac{\mathrm{KL}^{\mathrm{UB}}(\mathbb{Q}_h \| \mathbb{P}_{\mathrm{Sol}}) + \log(\xi(n)/\beta)}{n} \,,
\end{equation*}
and the bound on $\mathcal{R}_G(h)$ is obtained as the largest $p \in [\widehat{\mathcal{R}}_S(h),\,1]$ satisfying the inequality.
\end{proposition}

\begin{proposition}[Thiemann Certificate] \label{prop:thiemann_cert}
From the strongly quasiconvex bound of \citet{thiemann2017strongly},
\begin{equation*}
\mathcal{R}_G(h) \leq \inf_{\lambda \in (0, 2)} \left\{ \frac{\widehat{\mathcal{R}}_S(h)}{1 - \lambda/2} + \frac{\mathrm{KL}^{\mathrm{UB}}(\mathbb{Q}_h \| \mathbb{P}_{\mathrm{Sol}}) + \log(\xi(n)/\beta)}{n \lambda (1 - \lambda/2)} \right\} \,.
\end{equation*}
\end{proposition}

\begin{proposition}[Rivasplata Certificate] \label{prop:rivasplata_cert}
The empirical risk inversion bound \citep{rivasplata2019pac} avoids square roots of the complexity term in the dominant part of the bound:
\begin{equation*}
\mathcal{R}_G(h) \leq \widehat{\mathcal{R}}_S(h)+ \frac{C}{n} + \sqrt{2 \widehat{\mathcal{R}}_S(h)\frac{C}{n} + \left(\frac{C}{n}\right)^2} \,,
\end{equation*}
where $C = \mathrm{KL}^{\mathrm{UB}}(\mathbb{Q}_h \| \mathbb{P}_{\mathrm{Sol}}) + \log(\xi(n)/\beta)$.
\end{proposition}

\begin{figure}[ht!]
    \centering
    \includegraphics[width=0.6\linewidth]{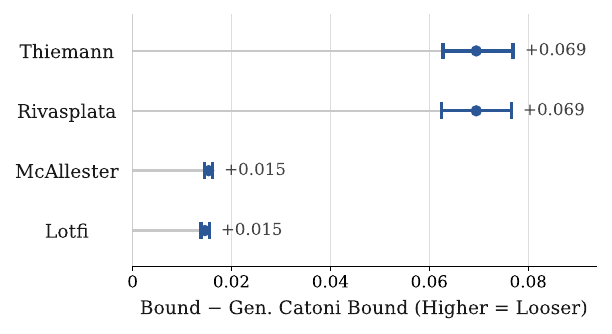}
    \setlength{\abovecaptionskip}{4pt}
    \caption{\textbf{Generalised Catoni yields the tightest PAC-Bayes bounds.} Mean difference in bound tightness relative to the generalised Catoni bound with 95\% bootstrap confidence intervals.}
    \vspace*{-10pt}
    \label{fig:bound-comparison}
\end{figure}

\Cref{fig:bound-comparison} evaluates \Cref{prop:lotfi_cert,prop:mcallester_cert,prop:thiemann_cert,prop:rivasplata_cert} in our setting. \Cref{prop:seeger_cert} is omitted because it is not numerically distinct: by the variational form $\mathrm{kl}(\widehat r \| p) = \sup_{\alpha>0}\{-\alpha \widehat r - \log(1-p(1-e^{-\alpha}))\}$ for $p\ge\widehat r$, the constraint $\mathrm{kl}(\widehat r \| p) \le C/n$ holds precisely when $p \le \Phi_{n\alpha}$ for every $\alpha>0$, so its largest root equals $\inf_{\lambda>0}\Phi_\lambda$. Under matching overhead \Cref{thm:catoni} is therefore the variational dual of \Cref{prop:seeger_cert}, evaluating the same value by minimisation in $\lambda$ rather than root-finding. The Hoeffding-type bounds of \Cref{prop:lotfi_cert,prop:mcallester_cert} are looser than the generalised Catoni bound by roughly $0.015$ on average, and the quasiconvex and inversion bounds of \Cref{prop:thiemann_cert,prop:rivasplata_cert} by roughly $0.069$, since both incur a penalty scaling as $\sqrt{\mathrm{KL}/n}$ rather than exploiting the exponential-moment structure Catoni-style bounds use. The choice of bound therefore accounts for a measurable part of certificate tightness, independently of the compression pipeline.

\clearpage
\section{Experimental setup}

We evaluate over four safety benchmarks, covering a range of unsafe categories.

\subsection{Dataset descriptions}

\textbf{Egida.}
The Egida dataset \citep{garcia2025efficient} focuses on evaluating the safety of agent responses to potentially harmful user queries. We use the Egida-HSafe configuration, drawing from the dataset's test split of 1,000 high-quality, manually labelled samples. As no pre-defined training partition is provided, we apply a fixed 80/20 train--test split and then drop entries whose prompt exceeds $5{,}000$ characters or whose response exceeds $10{,}000$ characters, yielding $792$ certification samples and $196$ test samples. Safety labels are derived by majority vote over five independent human annotators, with each annotator assigning a ``safe'' or ``unsafe'' label to the agent response. A sample is classified as unsafe if the majority vote is ``unsafe''. The taxonomy covers 9 hazard categories: Cybercrime, Non-Violent Crimes, Violent Crimes, Sexual Crimes and Erotic Content, Illegal Weapons and Substances, Hate and Harassment, Fake News and Misinformation, Dangerous Acts and Self-Harm, and Health.

\textbf{OpenAI Moderation.}
The OpenAI Moderation evaluation set \citep{markov2023holistic} provides labelled standalone user prompts drawn from interactions with the OpenAI API, with no accompanying agent response. Only the user prompt is evaluated for harm. The full dataset comprises 1,680 samples in a single split. As no pre-defined training partition is provided, we again apply a fixed 80/20 train--test split, yielding 1,344 training samples and 336 test samples. A sample is classified as unsafe if any of 8 binary moderation flags is positive. The taxonomy covers 8 hazard categories: Sexual Content, Hate Speech, Violence, Harassment, Self-Harm, Sexual Content Involving Minors, Hate Speech with Threats, and Graphic Violence.

\textbf{Aegis 2.0.}
Aegis 2.0 \citep{ghosh-etal-2025-aegis2} tasks the judge with jointly evaluating both the user prompt and the agent response for safety violations. We use the pre-defined train and test splits, filtering out entries marked as ``REDACTED'' and applying length constraints (prompts exceeding 5,000 characters and responses exceeding 10,000 characters) for more efficient training and inference. After filtering, this yields 28{,}984 training samples and 1{,}925 test samples. Each sample carries separate ``safe'' or ``unsafe'' labels for the prompt and response. A conversation is deemed unsafe if either label is ``unsafe''. The taxonomy covers 12 hazard categories: Hate/Identity Hate, Sexual Content, Suicide/Self-Harm, Violence, Guns/Illegal Weapons, Threats, PII/Privacy, Sexual Content (Minors), Criminal Planning, Harassment, Controlled Substances, and Profanity.

\textbf{BeaverTails.}
The BeaverTails dataset \citep{ji2023beavertails} assesses the safety of agent responses to user queries. Only the agent response is evaluated for harm. We use the pre-defined \texttt{330k\_train} and \texttt{330k\_test} splits, comprising 300{,}567 training samples and 33{,}396 test samples respectively. A sample is classified as unsafe if the \texttt{is\_safe} field evaluates to \texttt{False}. The taxonomy covers 14 hazards: Animal Abuse, Child Abuse, Controversial Topics/Politics, Discrimination/Stereotype, Drug Abuse/Weapons, Financial/Property Crime, Hate Speech, Misinformation, Non-Violent Unethical Behaviour, Privacy Violation, Self-Harm, Sexually Explicit, Terrorism/Organised Crime, and Violence/Incitement.

\subsection{Prompt templates}

\begin{figure}[ht!]
    \centering
    \includegraphics[width=1.0\linewidth]{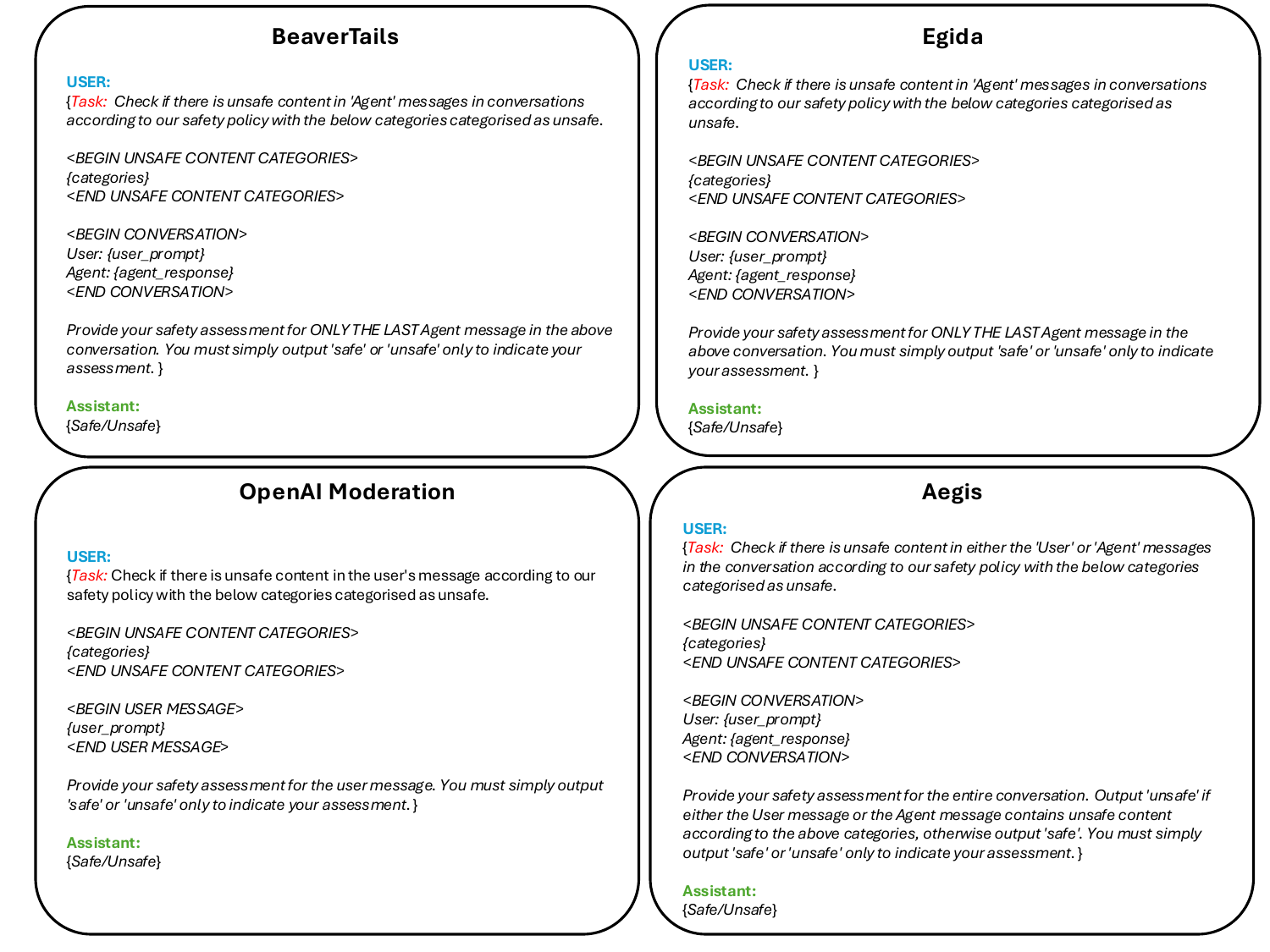}
    \caption{\textbf{Safety fine-tuning and evaluation prompts.} Prompts applied for both training and evaluation, shown separately for each of the four safety datasets and corresponding tasks considered in this work.}
    \label{fig:prompt-templates}
\end{figure}

All evaluations rely on a standardised prompt structure: \texttt{[Task Instruction] $\rightarrow$ [Safety Categories] $\rightarrow$ [Conversation] $\rightarrow$ [Assessment Instruction] $\rightarrow$ [Output]}, with the specific instructions varying by evaluation target. The templates for each dataset are given in \Cref{fig:prompt-templates}.

\subsection{Connection to the oversight framework}
\label{sec:oversight-connection}

Within the framework of \Cref{sec:judge-guarantees}, each benchmark instantiates a particular $\mathbb{P}_{\mathcal{D},G}$, where prompts are drawn from the benchmark's prompt distribution, responses are produced by the generator used to construct the benchmark, and the accompanying human or expert safety labels define $z^*$. \Cref{tab:generators} records the generator underlying each dataset, making explicit the distribution to which each certificate is relative.

On OpenAI Moderation only the user prompt is evaluated, so the response space is degenerate and $\mathbb{P}_G(\cdot \mid x)$ is a point mass on the empty response. The framework then reduces to the standard case in which risk is defined over the prompt distribution alone. On Aegis a conversation is labelled unsafe if either the prompt or the response violates the policy, so $z^*$ is determined by an unsafe prompt alone, before the generator acts. The certificate still covers the joint distribution, but part of the label does not depend on $y$. On Egida the responses are drawn from an ensemble of ten open-weight models, namely Phi-3-mini and Phi-3-medium, Yi-1.5-9B-Chat and Yi-1.5-34B-Chat, Llama-3.1-8B-Instruct and Llama-3.1-70B-Instruct, Qwen-2.5-7B-Instruct and Qwen-2.5-72B-Instruct, and Gemma-2-9b-it and Gemma-2-27b-it. A fixed mixture over generators is itself a fixed Markov kernel, so $G$ need not be a single network, and \Cref{cor:solomonoff_guard_base} requires only that $S$ be i.i.d.\ from the induced joint. The certificate is then relative to the mixture and does not transfer to an individual member, which would be a change of distribution (\Cref{app:scope}).

\begin{table}[t]
\centering
\caption{\textbf{Generators underlying each benchmark.} Each certificate is relative to the joint distribution induced by the corresponding generator and prompt distribution. On OpenAI Moderation only prompts are evaluated, so no generator is involved.}
\label{tab:generators}
\begin{tabular}{@{}llr@{}}
\toprule
Dataset & Response generator & $n$ \\
\midrule
Egida & Ensemble of ten open-weight LLMs & $792$ \\
OpenAI Moderation & None (prompts only) & $1{,}344$ \\
Aegis 2.0 & Mistral-7B-v0.1 & $28{,}984$ \\
BeaverTails & Alpaca-7B & $300{,}567$ \\
\bottomrule
\end{tabular}
\end{table}

\textbf{Why these distributions.~~} Each of these joints is a standard public benchmark on which safety judges are compared. A certificate computed on one is a guarantee about a quantity practitioners already report, making these joints the most useful and relevant to benchmark our methodology on, and \Cref{tab:generators} states the conditioning explicitly.

Transferring a certificate across a change of generator is out of scope (\Cref{app:scope}), and recomputing one for a fixed adaptor costs a single pass over a labelled sample from the new joint. We use these benchmarks because they are standard in the literature for training and evaluating safety judges, while our framework also accommodates changed generators by recomputing the certificate on labelled samples generated from the new distribution.

\subsection{Adaptor families and fine-tuning details}
\label{sec:peft-appendix}

\subsubsection{LoRA adaptor variants}
\label{sec:peft-variants}

\textbf{Low-rank adaptation (LoRA).}
LoRA \citep{hu2022lora} parameterises weight updates for a frozen, pre-trained linear transformation $W_0 \in \mathbb{R}^{d \times k}$ through a low-rank perturbation:
\begin{equation*}
    W = W_0 + BA,
\end{equation*}
where $B \in \mathbb{R}^{d \times r}$ and $A \in \mathbb{R}^{r \times k}$ are trainable with rank $r \leq \min(d, k)$.

\textbf{LoRA-XS.}
LoRA-XS \citep{balazy2024lora} improves on the parameter efficiency of LoRA by performing latent editing within the SVD-transformed space of the original weights:
\begin{equation*}
    W = W_0 + U \Sigma R V^\top,
\end{equation*}
where $U$, $\Sigma$ and $V$ are obtained via singular value decomposition of $W_0$ and remain frozen during training. By training only the small latent mapping matrix $R \in \mathbb{R}^{r \times r}$, LoRA-XS decouples the number of trainable parameters from the model's hidden dimensionality.

\textbf{LoRA-T.}
TinyLoRA (denoted LoRA-T here) \citep{morris2026learning} further compresses LoRA-XS by replacing $R$ with a low-dimensional trainable vector $\mathbf{v} \in \mathbb{R}^u$ projected through fixed matrices:
\begin{equation*}
    W = W_0 + U \Sigma \left( \sum_{i=1}^{u} v_i P_i \right) V^\top,
\end{equation*}
where the $P_i \in \mathbb{R}^{r \times r}$ are fixed random matrices and only $\mathbf{v}$ is updated. For further reduction, LoRA-T offers a weight-tied variant, \textit{LoRA-T Tied} (LoRA-TT in the main text), in which a single trainable vector $\mathbf{v}^{(\ell)} \in \mathbb{R}^u$ is shared across all targeted modules within transformer layer $\ell$ (e.g.\ all MLP and attention projections), while each module retains its own independent set of frozen projection matrices $P_i$. Our global-scale variant, LoRA-GT, shares a single quantisation scale across all layers rather than one per layer, and is defined in \Cref{sec:global-scale}.

These parameterisations produce compact updates whose complexity is measured by the description length of the adaptor after quantisation and compression.

\subsubsection{PEFT fine-tuning details}
\label{sec:peft_details}

All models are fine-tuned using the \texttt{SFTTrainer} from the Hugging Face TRL library, built on Transformers \citep{wolf2020transformers}, performing supervised fine-tuning (SFT) via PEFT on safe versus unsafe completion judgements according to each dataset's own taxonomy of hazards using a standard cross entropy loss.

\textbf{Shared hyperparameters across adaptor methods.~~} We use the AdamW optimiser \citep{loshchilov2018decoupled} with a cosine learning-rate schedule and a $3\%$ linear warm-up. All training is performed in bfloat16 precision. The maximum training sequence length is set per dataset to the 99th percentile of tokenised prompt lengths, rounded up to the nearest 64 tokens, so that fewer than $1\%$ of training examples are truncated (which avoids rare but extremely lengthy, and therefore computationally costly to deal with, examples). For inference we use dataset-appropriate context windows, $2048$ tokens for BeaverTails, $4096$ for OpenAI Moderation and $6000$ for Aegis and Egida.

\textbf{Adaptor-specific settings.~~} For each of the three adaptor families we sweep the following, with the three LoRA-T variants sharing identical training settings and differing only in tying and quantisation scope:

\begin{itemize}
    \item \textbf{LoRA:} ranks $r \in \{1, 2, 4\}$, with the scaling hyperparameter $\alpha$ set equal to the rank so that $\alpha/r = 1$. Learning rate $2 \times 10^{-4}$ with weight decay $0.01$, training for 4 epochs on the smaller datasets (Egida and OpenAI Moderation) and 1 epoch on the larger ones (Aegis and BeaverTails).

    \item \textbf{LoRA-XS:} the factors $U$, $\Sigma$ and $V$ are obtained by SVD of the base weights using 10 power iterations and frozen prior to training, so only the $r \times r$ latent mapping matrix $R$ is updated. We evaluate SVD ranks $r \in \{2, 4\}$, again setting $\alpha = r$. Learning rate $8 \times 10^{-4}$ with no weight decay, training for 6 epochs on the smaller datasets and 2 epochs on the larger ones.

    \item \textbf{LoRA-T:} as for LoRA-XS, the base factors are SVD-initialised in the same way as for LoRA-XS and frozen, with the trainable parameters restricted to the $u$-dimensional vector $\mathbf{v}$, which modifies the latent space via fixed random projections. We evaluate $(r, u) \in \{(2,1), (4,1), (4,2), (4,4)\}$, with $\alpha$ matched to the SVD rank. Learning rate $1 \times 10^{-3}$ with no weight decay, training for 6 epochs on the smaller datasets and 2 epochs on the larger ones.
\end{itemize}

For every adaptor family we target all linear layers within the transformer blocks, specifically all attention and MLP projections in each block.

\textbf{Model-specific hyperparameters and seeds.~~} Training batch sizes are adjusted by model scale to accommodate GPU memory constraints, $8$ for Qwen3Guard-Gen-0.6B and Llama-Guard-3-1B, and $4$ for the 4B and 8B models, in each case across all four datasets. Every configuration is trained with a common seed, held fixed across runs, which is itself stored in the certificate and charged to $|h|$ as described in \Cref{sec:quantisation_details}.

\subsection{Computational resources}
\label{app:compute}\label{sec:compute_resources}

Fine-tuning for Llama-Guard-3-1B and Qwen3Guard-Gen-0.6B was performed on a single NVIDIA L40S GPU with 48~GB of VRAM, as was fine-tuning for Qwen3Guard-Gen-4B. The 8B models, Llama-Guard-3-8B and Qwen3Guard-Gen-8B, used a single NVIDIA H100 (80~GB VRAM). The sweep is dominated by supervised fine-tuning on BeaverTails.

Certification is negligible by comparison. It is computed post hoc on the sample used for fine-tuning, so no data is held out and no retraining is needed. For a trained adaptor it requires quantisation and lossless encoding, followed by one forward pass over $S$ for the empirical risk. This contrasts with validation-based guarantees, which give up training data (\Cref{app:validation-comparison}), and with stochastic-posterior certificates, which need repeated sampling to approximate Gibbs risks (\Cref{app:detpost}).

\section{Quantisation and compression}
\label{sec:quantisation_details}

To minimise the complexity penalty in our certificates, we then pass the fine-tuned trainable adaptor weights to a post-training quantisation and compression pipeline, implemented using the \texttt{llm-compressor} library \citep{llmcompressor2024}. All adaptors first undergo deterministic Hadamard incoherence processing, as in QuIP \citep{tseng2024quip}. LoRA and LoRA-XS then use GPTQ \citep{frantar2022gptq}, while LoRA-TN, LoRA-TT and LoRA-GT use nearest-integer rounding.

\subsection{Quantisation pipeline}

\textbf{Hadamard incoherence processing.~~} To mitigate outliers and reduce quantisation error, we apply deterministic, normalised Hadamard rotations \citep{tseng2024quip,sylvester1867lx}. Using \Cref{sec:quant-compression}, let $\theta\in\mathbb{R}^p$ contain the entries of one trainable adaptor tensor, and let $\operatorname{mat}(\theta)\in\mathbb{R}^{d_\theta\times k_\theta}$ denote its matrix form, with $p=d_\theta k_\theta$. The shape $(d_\theta,k_\theta)$ is $(r,k)$ for $A$, $(d,r)$ for $B$, $(r,r)$ for $R$, and $(1,u)$ for $\mathbf{v}^\top$ (\Cref{sec:peft-variants}). With orthogonal Hadamard matrices $H_{\mathrm{in}}\in\mathbb{R}^{k_\theta\times k_\theta}$ and $H_{\mathrm{out}}\in\mathbb{R}^{d_\theta\times d_\theta}$, the rotation and reconstruction are
\begin{equation*}
\begin{aligned}
\operatorname{mat}(\widetilde{\theta}) &= H_{\mathrm{out}}\operatorname{mat}(\theta)H_{\mathrm{in}}^\top, \\
\operatorname{mat}(\widehat{\theta}) &= H_{\mathrm{out}}^\top\operatorname{mat}\bigl(d_b(z,s)\bigr)H_{\mathrm{in}}.
\end{aligned}
\end{equation*}
Here $\widetilde{\theta},\widehat{\theta}\in\mathbb{R}^p$, $z\in\mathcal{C}_b^p$ and $s\in\mathcal{S}$, with $d_b$ as defined in \Cref{sec:quant-compression}. For LoRA-T, $H_{\mathrm{out}}=[1]$, so only the input-side rotation is applied, equivalently $\widetilde{\mathbf{v}}=H_{\mathrm{in}}\mathbf{v}\in\mathbb{R}^u$. The reconstructed parameters are restored in place after inverting the rotations.

\textbf{GPTQ calibration and configuration.~~} We use GPTQ \citep{frantar2022gptq} for LoRA and LoRA-XS to reduce output reconstruction error at each linear adaptor operation. Let $X\in\mathbb{R}^{k_\theta\times m_{\mathrm{cal}}}$ contain $m_{\mathrm{cal}}\in\mathbb{N}$ calibration input vectors, and let $\widetilde{X}=H_{\mathrm{in}}X\in\mathbb{R}^{k_\theta\times m_{\mathrm{cal}}}$ denote their rotated form. With these inputs and scales $s\in\mathcal{S}$ fixed, GPTQ approximately solves
\begin{equation*}
\min_{z\in\mathcal{C}_b^p}\;\bigl\|\operatorname{mat}\bigl(\widetilde{\theta}-d_b(z,s)\bigr)\widetilde{X}\bigr\|_F^2,
\end{equation*}
where here $\|\cdot\|_F$ denotes the Frobenius norm. Calibration uses the fine-tuning subsample with the same tokenisation and prompt formatting. We use a damping factor of $0.05$ to stabilise Hessian inversion. The block size is $r$ for LoRA and LoRA-XS. Quantisation is symmetric at $b$-bit precision.

\textbf{Nearest-integer rounding and scale choice.~~} LoRA-TN, LoRA-TT and LoRA-GT use nearest-integer rounding. LoRA-TT applies the clipping-free scale $s^*$ of \Cref{prop:scale} per layer to $\mathbf{v}^{(\ell)}$, so that $s \in \mathbb{R}_{>0}^L$. LoRA-GT applies one scale to the concatenated $\mathbf{v} \in \mathbb{R}^{Lu}$, so that $s \in \mathbb{R}_{>0}$, chosen by the search of \Cref{sec:global-scale} at $b \le 4$ and equal to $s^*$ at $b \ge 8$, where scale search offers negligible improvement (\Cref{app:scale-search-bits}).

\subsection{Scale search by bit-width}
\label{app:scale-search-bits}

The search of \Cref{sec:global-scale} runs at $b \le 4$, and $s^*$ is used at $b \ge 8$. Lowering the scale below $s^*$ reduces the rounding-error bound $s/2$ on unclipped coordinates, but can increase clipping error. Clipping occurs when $|v_i|>s(q_{\max}+\tfrac12)$ under the rounding convention of \Cref{prop:scale}. At high $b$ the rounding error at $s^*$ is already small. Write $E(s) = \|\mathbf{v} - \hat{\mathbf{v}}(s)\|_2^2$ for the squared reconstruction error at scale $s$ and, for $c \in (0,1]$,
\begin{equation*}
K(c) = \sum_{i \,:\, |v_i| > c\,\|\mathbf{v}\|_\infty} \bigl(|v_i| - c\,\|\mathbf{v}\|_\infty\bigr)^2
\end{equation*}
for the sum of squared excesses above the threshold $c\,\|\mathbf{v}\|_\infty$.

\begin{proposition}[Scales below $s^*$]\label{prop:scale-search-bits}
Fix $b \in \mathbb{N}$ and nonzero $\mathbf{v} \in \mathbb{R}^{Lu}$, and let $s^*$ be as in \Cref{prop:scale}. Then
\begin{equation*}
E(s^*) \le \frac{Lu\,\|\mathbf{v}\|_\infty^2}{(2q_{\max}+1)^2}
\qquad\text{and}\qquad
E(c\,s^*) \ge K(c) \quad \text{for every } c \in (0,1].
\end{equation*}
Hence $E(c\,s^*) < E(s^*)$ only if $K(c) < Lu\,\|\mathbf{v}\|_\infty^2/(2q_{\max}+1)^2$, and so only if
\begin{equation*}
c \;\ge\; 1 - \frac{\sqrt{Lu}}{2q_{\max}+1}.
\end{equation*}
\end{proposition}

\begin{proof}
By \Cref{prop:scale}, $s^*$ is clipping-free and $|v_i - \hat v_i(s^*)| \le s^*/2$ for every $i$, so $E(s^*) \le Lu\,(s^*)^2/4 = Lu\,\|\mathbf{v}\|_\infty^2/(2q_{\max}+1)^2$. Now fix $s = c\,s^*$. Since $s(q_{\max}+\tfrac12)=c\,\|\mathbf{v}\|_\infty$, every coordinate with $|v_i|>c\,\|\mathbf{v}\|_\infty$ is clipped to $\pm q_{\max}$. Its reconstruction error satisfies
\begin{equation*}
|v_i-\hat v_i(s)|=|v_i|-s\,q_{\max}\ge |v_i|-c\,\|\mathbf{v}\|_\infty.
\end{equation*}
Squaring and summing over these coordinates gives $E(s)\ge K(c)$. The last claim follows from $K(c)\ge(1-c)^2\,\|\mathbf{v}\|_\infty^2$, the contribution of a coordinate attaining $\|\mathbf{v}\|_\infty$.
\end{proof}

At $b = 2$ the lower bound on $c$ is negative, so the proposition places no restriction. For the vectors here $Lu \le 144$, so an improving scale lies in $[0.2\,s^*, s^*]$ at $b = 4$, in $[0.95\,s^*, s^*]$ at $b = 8$ and in $[0.9998\,s^*, s^*]$ at $b = 16$. The scale is stored in bfloat16, with relative spacing between $2^{-8}$ and $2^{-7}$, so the window narrows to a handful of representable scales at $8$ bits and to those within one bfloat16 step of $s^*$ at $16$. The proposition concerns the ideal scale $s^*$. Empirical comparisons use the smallest bfloat16 value at or above $s^*$ as the reference, with both scales stored in bfloat16.

\Cref{tab:scale-search-bits} reports the search on the $60$ fine-tuned LoRA-GT vectors of the five main models at $2$, $3$, $4$ and $8$ bits. The search lowers the relative error $\|\mathbf{v} - \hat{\mathbf{v}}(s)\|_2 / \|\mathbf{v}\|_2$ by $11\%$ at $2$ bits, $7\%$ at $3$ and $3\%$ at $4$, clipping one to four coordinates. At $8$ bits, the mean relative error at the reference scale is below $0.5\%$, and scale search reduces it by only about $2\times10^{-5}$. This small absolute gain motivates using $s^*$ at higher bit-widths. The effects on risk and certificate tightness are reported separately in \Cref{tab:scale-intervention}.

\begin{table}[h]
\centering
\caption{\textbf{Reconstruction error of LoRA-GT at the no-clipping scale and the searched scale.} Relative $\ell_2$ error of the dequantised vector, mean over the $60$ fine-tuned vectors (five models, four benchmarks, $u \in \{1,2,4\}$), with the mean number of coordinates the searched scale clips. Both columns use bfloat16-representable scales, the precision the decoder stores. The search lowers the error materially only at $b \le 4$.}
\label{tab:scale-search-bits}
\small
\begin{tabular}{lcccc}
\toprule
$b$ & Error at $s^*$ & Error at searched $s$ & Reduction & Coordinates clipped \\
\midrule
$2$ & $0.409$ & $0.365$ & $11\%$ & $4.0$ \\
$3$ & $0.177$ & $0.165$ & $7\%$ & $1.5$ \\
$4$ & $0.081$ & $0.079$ & $3\%$ & $1.0$ \\
$8$ & $0.00489$ & $0.00487$ & $0.3\%$ & $0.1$ \\
\bottomrule
\end{tabular}
\end{table}

\subsection{Compression certificate construction}
\label{app:certconstruct}
For a PAC-Bayes certificate to remain valid, the prior must be constructed independently of the training data. Under the Solomonoff prior, the KL divergence between this data-independent prior and the deterministic posterior is upper-bounded by an expression involving the description length (\Cref{prop:kl-ub}), which we measure empirically as the size of an encoded file storing the trained adaptor. To preserve validity, this file must encode all information dependent on the training and certification sample $S$ that is needed to reconstruct the posterior hypothesis.

\textbf{Components included in the certificate.~~} The encoded bit-stream contains:
\begin{enumerate}
    \item \textbf{Packed quantised weights.} The trainable parameters, symmetrically quantised to the target bit-width and packed at exactly $b$ bits per value into a contiguous bitstream.
    \item \textbf{Quantisation scales.} The scale factors needed to dequantise, retaining their original precision (e.g.\ \texttt{bfloat16}). These are stored per tensor for LoRA, LoRA-XS and LoRA-TN, per layer for LoRA-TT, and as a single scalar for LoRA-GT, which is the metadata reduction analysed in \Cref{sec:global-scale}.
    \item \textbf{Random seed.} The scalar seed initialising all stochastic components of the pipeline (SVD initialisation of the frozen factors, generation of the LoRA-T projection matrices, data shuffling). Its Elias $\delta$ code length is included in $|h|$.
\end{enumerate}

\textbf{Components excluded from the certificate.~~} The following are excluded because they are either data-independent, and hence part of the fixed decoder $M$, or unnecessary for reconstruction:
\begin{enumerate}
    \item \textbf{Backbone weights.} The pre-trained weights of the frozen safety oversight model are shared between prior and posterior, and are fixed before $S$ is observed (\Cref{app:scope}).
    \item \textbf{Frozen factors and projection matrices.} For LoRA-XS and LoRA-T the factors $U$, $\Sigma$ and $V$ are obtained by SVD of the base weights and frozen before training. For LoRA-T the projection matrices $P_i$ are also excluded, since they are reconstructible exactly from the stored seed.
    \item \textbf{Hadamard rotation matrices.} The Hadamard transforms $H_{\mathrm{in}}$ and $H_{\mathrm{out}}$ are generated deterministically from the matrix dimensions, carry no information derived from $S$, and form part of the fixed decoding scheme.
    \item \textbf{Zero-points and group indices.} Since quantisation is symmetric and without activation reordering, zero-points are identically zero and no data-dependent column permutation exists, so neither requires any bits.
\end{enumerate}

\textbf{Configuration selection overhead.~~} Several results in the main text report the tightest certificate obtained within an adaptor family across a finite set of configurations. Such models may be viewed as elements of a common enlarged hypothesis class, and under the coding interpretation of the Solomonoff prior the selected settings, such as adaptor rank and bit-width, form part of the hypothesis description. Recording which of $K$ configurations was chosen takes $\lceil \log_2 K \rceil$ bits (\Cref{app:no-leakage}). Alternatively, a union bound over $K$ configuration-specific bounds, allocating failure probability $\beta/K$ to each, adds $\log K$ to the confidence term, equivalent to $\log_2 K$ bits without rounding. The candidate set depends on which selection is being reported. Where \Cref{fig:main-results} and \Crefrange{tab:pipeline-egida}{tab:pipeline-beavertails} report the tightest certificate within a family, $K$ is the full family grid, $3$ ranks $\times$ $2$ bit-widths for LoRA ($K = 6$), $2 \times 2$ for LoRA-XS ($K = 4$), and $2$ bit-widths across the $4$ $(r,u)$ settings of LoRA-TN and LoRA-TT and the $3$ of LoRA-GT ($K \le 22$), adding at most $5$ selection bits. Because the overhead grows only logarithmically in $K$, it is negligible relative to the adaptor description lengths considered here, and even selecting among a thousand configurations would add at most $10$ bits.

\textbf{Compression.}
We apply lossless Brotli compression \citep{google_brotli_v020} at its maximum setting (\texttt{quality=11}) to the bit-stream containing the components above, and retain the shorter of the raw and compressed representations. A one-bit flag identifies the choice, giving
\begin{equation*}
|h| = 1 + \min\{L_{\mathrm{raw}},L_{\mathrm{Brotli}}\},
\end{equation*}
where both lengths are in bits. This avoids Brotli's format overhead when it exceeds the compression savings for small adaptors. Both decoding branches are fixed independently of $S$, and the Elias length encoding of \Cref{prop:kl-ub} makes the complete description prefix-free. \Cref{app:dl-compressor} compares the compressors.

\subsection{Full pipeline}
\label{sec:pipeline}

\begin{figure}[ht!]
    \centering
    \vspace{-0.5em}
    \includegraphics[width=1.0\linewidth]{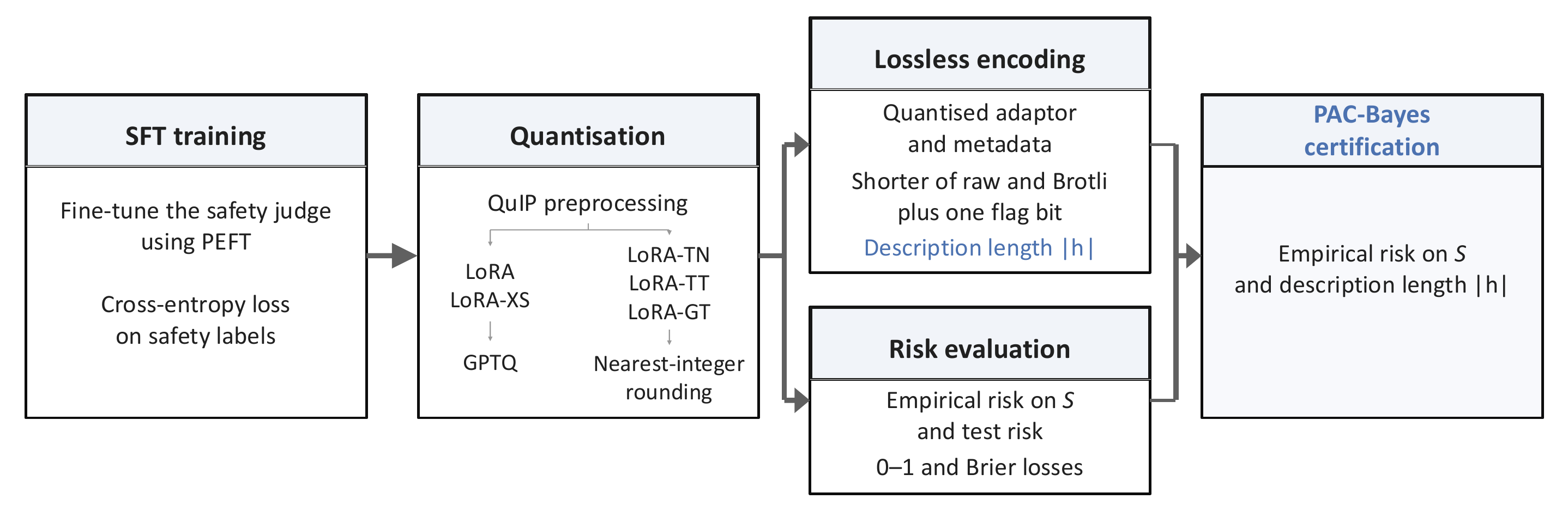}
    \vspace{-0.5em}
    \caption{\textbf{Training and certification pipeline.} Models are trained via SFT, quantised using Hadamard incoherence processing with either GPTQ (LoRA, LoRA-XS) or nearest-integer rounding (LoRA-TN, LoRA-TT, LoRA-GT), encoded using the shorter of raw and Brotli with a one-bit flag, and evaluated using empirical test risk and PAC-Bayes certificates.}
    \label{fig:pipeline}
    \vspace{-0.5em}
\end{figure}

\Cref{fig:pipeline} summarises the training and certification pipeline. The training set serves three purposes, supervised fine-tuning of the oversight model, calibration for post-training quantisation, and computation of the empirical risk entering the certificate. The quantised adaptor is then losslessly compressed to obtain its description length $|h|$, and evaluation reports both empirical test risk and the resulting PAC-Bayes certificate. No held-out split is introduced at any stage (\Cref{app:no-leakage}).

\newpage
\section{Additional results and ablations}

\FloatBarrier
\subsection{Complete per-model results}
\label{app:detailed-results}

We report the full per-model results from \cref{sec:experiments} for all datasets and both losses. \Crefrange{tab:pipeline-egida}{tab:pipeline-beavertails} use the same selection and adaptor families as \Cref{fig:main-results}, whose LoRA-T entries are the LoRA-GT rows (the other LoRA-T variants are compared in \Cref{app:tinylora-variants}). For each adaptor family and base model, the reported configuration is the one with the tightest Catoni bound ($\beta = 0.05$) over the $(r,u)$ settings and bit-widths of the family. Every LoRA bound is vacuous, so LoRA is reported at its lowest test risk instead. Each entry gives the test risk of that adaptor before (FT) and after (Q) quantisation and its certificate including the selection charge (\Cref{app:certconstruct}). Within each table, \underline{underlining} marks the lowest test risk per model and loss and \textbf{bold} marks the tightest bound.

\begin{table*}[ht!]
\centering
\caption{\textbf{Complete results on Egida.} Test risk before (FT) and after (Q) quantisation, with Catoni bounds including the selection charge ($\beta=0.05$). Selected configurations appear below each row. Underlining marks the lowest test risk and bold the tightest bound per model and loss.}
\label{tab:pipeline-egida}
\setlength{\tabcolsep}{4pt}
\renewcommand{\arraystretch}{1.10}
\resizebox{\textwidth}{!}{%
\begin{tabular}{l rrr @{\hspace{12pt}} rrr @{\hspace{12pt}} rrr @{\hspace{12pt}} rrr @{\hspace{12pt}} rrr}
\toprule
 & \multicolumn{3}{c}{\shortstack{Qwen3Guard\\0.6B}} & \multicolumn{3}{c}{\shortstack{Llama-Guard-3\\1B}} & \multicolumn{3}{c}{\shortstack{Qwen3Guard\\4B}} & \multicolumn{3}{c}{\shortstack{Llama-Guard-3\\8B}} & \multicolumn{3}{c}{\shortstack{Qwen3Guard\\8B}} \\
\cmidrule(lr){2-4}\cmidrule(lr){5-7}\cmidrule(lr){8-10}\cmidrule(lr){11-13}\cmidrule(lr){14-16}
\textit{0--1 loss} & FT & Q & Bound & FT & Q & Bound & FT & Q & Bound & FT & Q & Bound & FT & Q & Bound \\
\cmidrule(lr){1-16}
LoRA & .092 & \underline{.087} & 1.000 & \underline{.112} & .112 & 1.000 & \underline{.087} & .087 & 1.000 & \underline{.112} & .112 & 1.000 & .102 & \underline{.097} & 1.000 \\
 & \multicolumn{3}{c}{\scriptsize(r=2, 2-bit)} & \multicolumn{3}{c}{\scriptsize(r=1, 4-bit)} & \multicolumn{3}{c}{\scriptsize(r=2, 4-bit)} & \multicolumn{3}{c}{\scriptsize(r=2, 4-bit)} & \multicolumn{3}{c}{\scriptsize(r=2, 4-bit)} \\
\addlinespace[3pt]
LoRA-XS & .153 & .158 & .991 & .158 & .163 & .961 & .158 & .158 & .995 & .133 & .138 & .994 & .138 & .143 & .996 \\
 & \multicolumn{3}{c}{\scriptsize(r=2, 2-bit)} & \multicolumn{3}{c}{\scriptsize(r=2, 2-bit)} & \multicolumn{3}{c}{\scriptsize(r=2, 2-bit)} & \multicolumn{3}{c}{\scriptsize(r=2, 2-bit)} & \multicolumn{3}{c}{\scriptsize(r=2, 2-bit)} \\
\addlinespace[3pt]
LoRA-GT & .173 & .168 & \textbf{.367} & .194 & .204 & \textbf{.395} & .179 & .173 & \textbf{.361} & .148 & .143 & \textbf{.382} & .153 & .158 & \textbf{.360} \\
 & \multicolumn{3}{c}{\scriptsize(u=1, 2-bit)} & \multicolumn{3}{c}{\scriptsize(u=1, 2-bit)} & \multicolumn{3}{c}{\scriptsize(u=1, 2-bit)} & \multicolumn{3}{c}{\scriptsize(u=1, 2-bit)} & \multicolumn{3}{c}{\scriptsize(u=1, 2-bit)} \\
\midrule
\textit{Brier loss} & FT & Q & Bound & FT & Q & Bound & FT & Q & Bound & FT & Q & Bound & FT & Q & Bound \\
\cmidrule(lr){1-16}
LoRA & .083 & \underline{.080} & 1.000 & .085 & \underline{.084} & 1.000 & .085 & \underline{.079} & 1.000 & .102 & \underline{.101} & 1.000 & \underline{.074} & .078 & 1.000 \\
 & \multicolumn{3}{c}{\scriptsize(r=2, 2-bit)} & \multicolumn{3}{c}{\scriptsize(r=1, 4-bit)} & \multicolumn{3}{c}{\scriptsize(r=4, 2-bit)} & \multicolumn{3}{c}{\scriptsize(r=2, 2-bit)} & \multicolumn{3}{c}{\scriptsize(r=1, 2-bit)} \\
\addlinespace[3pt]
LoRA-XS & .109 & .111 & .986 & .118 & .120 & .944 & .124 & .128 & .994 & .103 & .104 & .991 & .122 & .117 & .994 \\
 & \multicolumn{3}{c}{\scriptsize(r=2, 2-bit)} & \multicolumn{3}{c}{\scriptsize(r=2, 2-bit)} & \multicolumn{3}{c}{\scriptsize(r=2, 2-bit)} & \multicolumn{3}{c}{\scriptsize(r=2, 2-bit)} & \multicolumn{3}{c}{\scriptsize(r=2, 2-bit)} \\
\addlinespace[3pt]
LoRA-GT & .123 & .125 & \textbf{.307} & .137 & .141 & \textbf{.314} & .142 & .140 & \textbf{.337} & .111 & .118 & \textbf{.322} & .129 & .132 & \textbf{.320} \\
 & \multicolumn{3}{c}{\scriptsize(u=1, 2-bit)} & \multicolumn{3}{c}{\scriptsize(u=1, 2-bit)} & \multicolumn{3}{c}{\scriptsize(u=1, 2-bit)} & \multicolumn{3}{c}{\scriptsize(u=1, 2-bit)} & \multicolumn{3}{c}{\scriptsize(u=1, 2-bit)} \\
\bottomrule
\end{tabular}
}
\end{table*}

\begin{table*}[ht!]
\centering
\caption{\textbf{Complete results on OpenAI Moderation.} Test risk before (FT) and after (Q) quantisation, with Catoni bounds including the selection charge ($\beta=0.05$). Selected configurations appear below each row. Underlining marks the lowest test risk and bold the tightest bound per model and loss.}
\label{tab:pipeline-openai_moderation}
\setlength{\tabcolsep}{4pt}
\renewcommand{\arraystretch}{1.10}
\resizebox{\textwidth}{!}{%
\begin{tabular}{l rrr @{\hspace{12pt}} rrr @{\hspace{12pt}} rrr @{\hspace{12pt}} rrr @{\hspace{12pt}} rrr}
\toprule
 & \multicolumn{3}{c}{\shortstack{Qwen3Guard\\0.6B}} & \multicolumn{3}{c}{\shortstack{Llama-Guard-3\\1B}} & \multicolumn{3}{c}{\shortstack{Qwen3Guard\\4B}} & \multicolumn{3}{c}{\shortstack{Llama-Guard-3\\8B}} & \multicolumn{3}{c}{\shortstack{Qwen3Guard\\8B}} \\
\cmidrule(lr){2-4}\cmidrule(lr){5-7}\cmidrule(lr){8-10}\cmidrule(lr){11-13}\cmidrule(lr){14-16}
\textit{0--1 loss} & FT & Q & Bound & FT & Q & Bound & FT & Q & Bound & FT & Q & Bound & FT & Q & Bound \\
\cmidrule(lr){1-16}
LoRA & \underline{.116} & .116 & 1.000 & .113 & \underline{.110} & 1.000 & \underline{.098} & .098 & 1.000 & .095 & \underline{.080} & 1.000 & \underline{.104} & .107 & 1.000 \\
 & \multicolumn{3}{c}{\scriptsize(r=2, 4-bit)} & \multicolumn{3}{c}{\scriptsize(r=1, 2-bit)} & \multicolumn{3}{c}{\scriptsize(r=1, 2-bit)} & \multicolumn{3}{c}{\scriptsize(r=1, 2-bit)} & \multicolumn{3}{c}{\scriptsize(r=1, 2-bit)} \\
\addlinespace[3pt]
LoRA-XS & .155 & .146 & .933 & .238 & .250 & .885 & .134 & .134 & .964 & .101 & .104 & .939 & .122 & .137 & .965 \\
 & \multicolumn{3}{c}{\scriptsize(r=2, 2-bit)} & \multicolumn{3}{c}{\scriptsize(r=2, 2-bit)} & \multicolumn{3}{c}{\scriptsize(r=2, 2-bit)} & \multicolumn{3}{c}{\scriptsize(r=2, 2-bit)} & \multicolumn{3}{c}{\scriptsize(r=2, 2-bit)} \\
\addlinespace[3pt]
LoRA-GT & .158 & .158 & \textbf{.258} & .182 & .202 & \textbf{.305} & .149 & .149 & \textbf{.298} & .131 & .134 & \textbf{.259} & .158 & .143 & \textbf{.265} \\
 & \multicolumn{3}{c}{\scriptsize(u=1, 2-bit)} & \multicolumn{3}{c}{\scriptsize(u=1, 2-bit)} & \multicolumn{3}{c}{\scriptsize(u=2, 2-bit)} & \multicolumn{3}{c}{\scriptsize(u=1, 2-bit)} & \multicolumn{3}{c}{\scriptsize(u=1, 2-bit)} \\
\midrule
\textit{Brier loss} & FT & Q & Bound & FT & Q & Bound & FT & Q & Bound & FT & Q & Bound & FT & Q & Bound \\
\cmidrule(lr){1-16}
LoRA & \underline{.086} & .086 & 1.000 & .097 & \underline{.094} & 1.000 & .076 & \underline{.074} & 1.000 & .078 & \underline{.072} & 1.000 & .080 & \underline{.077} & 1.000 \\
 & \multicolumn{3}{c}{\scriptsize(r=1, 4-bit)} & \multicolumn{3}{c}{\scriptsize(r=1, 2-bit)} & \multicolumn{3}{c}{\scriptsize(r=1, 2-bit)} & \multicolumn{3}{c}{\scriptsize(r=4, 2-bit)} & \multicolumn{3}{c}{\scriptsize(r=1, 2-bit)} \\
\addlinespace[3pt]
LoRA-XS & .103 & .105 & .924 & .166 & .168 & .851 & .096 & .099 & .958 & .078 & .079 & .931 & .090 & .091 & .957 \\
 & \multicolumn{3}{c}{\scriptsize(r=2, 2-bit)} & \multicolumn{3}{c}{\scriptsize(r=2, 2-bit)} & \multicolumn{3}{c}{\scriptsize(r=2, 2-bit)} & \multicolumn{3}{c}{\scriptsize(r=2, 2-bit)} & \multicolumn{3}{c}{\scriptsize(r=2, 2-bit)} \\
\addlinespace[3pt]
LoRA-GT & .111 & .112 & \textbf{.217} & .137 & .142 & \textbf{.240} & .103 & .109 & \textbf{.267} & .099 & .104 & \textbf{.220} & .120 & .110 & \textbf{.236} \\
 & \multicolumn{3}{c}{\scriptsize(u=1, 2-bit)} & \multicolumn{3}{c}{\scriptsize(u=1, 2-bit)} & \multicolumn{3}{c}{\scriptsize(u=2, 2-bit)} & \multicolumn{3}{c}{\scriptsize(u=1, 2-bit)} & \multicolumn{3}{c}{\scriptsize(u=1, 2-bit)} \\
\bottomrule
\end{tabular}
}
\end{table*}

\begin{table*}[ht!]
\centering
\caption{\textbf{Complete results on Aegis.} Test risk before (FT) and after (Q) quantisation, with Catoni bounds including the selection charge ($\beta=0.05$). Selected configurations appear below each row. Underlining marks the lowest test risk and bold the tightest bound per model and loss.}
\label{tab:pipeline-aegis}
\setlength{\tabcolsep}{4pt}
\renewcommand{\arraystretch}{1.10}
\resizebox{\textwidth}{!}{%
\begin{tabular}{l rrr @{\hspace{12pt}} rrr @{\hspace{12pt}} rrr @{\hspace{12pt}} rrr @{\hspace{12pt}} rrr}
\toprule
 & \multicolumn{3}{c}{\shortstack{Qwen3Guard\\0.6B}} & \multicolumn{3}{c}{\shortstack{Llama-Guard-3\\1B}} & \multicolumn{3}{c}{\shortstack{Qwen3Guard\\4B}} & \multicolumn{3}{c}{\shortstack{Llama-Guard-3\\8B}} & \multicolumn{3}{c}{\shortstack{Qwen3Guard\\8B}} \\
\cmidrule(lr){2-4}\cmidrule(lr){5-7}\cmidrule(lr){8-10}\cmidrule(lr){11-13}\cmidrule(lr){14-16}
\textit{0--1 loss} & FT & Q & Bound & FT & Q & Bound & FT & Q & Bound & FT & Q & Bound & FT & Q & Bound \\
\cmidrule(lr){1-16}
LoRA & \underline{.117} & .118 & 1.000 & \underline{.118} & .118 & 1.000 & .114 & \underline{.111} & 1.000 & .109 & \underline{.106} & 1.000 & .113 & \underline{.112} & 1.000 \\
 & \multicolumn{3}{c}{\scriptsize(r=4, 4-bit)} & \multicolumn{3}{c}{\scriptsize(r=4, 4-bit)} & \multicolumn{3}{c}{\scriptsize(r=1, 4-bit)} & \multicolumn{3}{c}{\scriptsize(r=2, 2-bit)} & \multicolumn{3}{c}{\scriptsize(r=4, 4-bit)} \\
\addlinespace[3pt]
LoRA-XS & .150 & .155 & .315 & .178 & .185 & .382 & .136 & .138 & .397 & .166 & .178 & .395 & .146 & .149 & .393 \\
 & \multicolumn{3}{c}{\scriptsize(r=2, 2-bit)} & \multicolumn{3}{c}{\scriptsize(r=4, 2-bit)} & \multicolumn{3}{c}{\scriptsize(r=2, 4-bit)} & \multicolumn{3}{c}{\scriptsize(r=2, 2-bit)} & \multicolumn{3}{c}{\scriptsize(r=2, 4-bit)} \\
\addlinespace[3pt]
LoRA-GT & .169 & .177 & \textbf{.186} & .268 & .282 & \textbf{.322} & .142 & .139 & \textbf{.227} & .174 & .176 & \textbf{.235} & .146 & .151 & \textbf{.178} \\
 & \multicolumn{3}{c}{\scriptsize(u=2, 4-bit)} & \multicolumn{3}{c}{\scriptsize(u=4, 4-bit)} & \multicolumn{3}{c}{\scriptsize(u=4, 4-bit)} & \multicolumn{3}{c}{\scriptsize(u=4, 4-bit)} & \multicolumn{3}{c}{\scriptsize(u=2, 4-bit)} \\
\midrule
\textit{Brier loss} & FT & Q & Bound & FT & Q & Bound & FT & Q & Bound & FT & Q & Bound & FT & Q & Bound \\
\cmidrule(lr){1-16}
LoRA & \underline{.087} & .087 & 1.000 & \underline{.086} & .086 & 1.000 & \underline{.081} & .082 & 1.000 & \underline{.080} & .081 & 1.000 & \underline{.081} & .082 & 1.000 \\
 & \multicolumn{3}{c}{\scriptsize(r=4, 4-bit)} & \multicolumn{3}{c}{\scriptsize(r=4, 2-bit)} & \multicolumn{3}{c}{\scriptsize(r=4, 4-bit)} & \multicolumn{3}{c}{\scriptsize(r=2, 2-bit)} & \multicolumn{3}{c}{\scriptsize(r=4, 4-bit)} \\
\addlinespace[3pt]
LoRA-XS & .108 & .109 & .268 & .155 & .170 & .309 & .112 & .112 & .362 & .117 & .123 & .328 & .117 & .118 & .358 \\
 & \multicolumn{3}{c}{\scriptsize(r=2, 2-bit)} & \multicolumn{3}{c}{\scriptsize(r=2, 2-bit)} & \multicolumn{3}{c}{\scriptsize(r=2, 4-bit)} & \multicolumn{3}{c}{\scriptsize(r=2, 2-bit)} & \multicolumn{3}{c}{\scriptsize(r=2, 4-bit)} \\
\addlinespace[3pt]
LoRA-GT & .118 & .120 & \textbf{.142} & .180 & .182 & \textbf{.224} & .118 & .121 & \textbf{.189} & .124 & .123 & \textbf{.183} & .114 & .118 & \textbf{.148} \\
 & \multicolumn{3}{c}{\scriptsize(u=2, 2-bit)} & \multicolumn{3}{c}{\scriptsize(u=2, 4-bit)} & \multicolumn{3}{c}{\scriptsize(u=2, 4-bit)} & \multicolumn{3}{c}{\scriptsize(u=4, 4-bit)} & \multicolumn{3}{c}{\scriptsize(u=2, 4-bit)} \\
\bottomrule
\end{tabular}
}
\end{table*}

\begin{table*}[ht!]
\centering
\caption{\textbf{Complete results on BeaverTails.} Test risk before (FT) and after (Q) quantisation, with Catoni bounds including the selection charge ($\beta=0.05$). Selected configurations appear below each row. Underlining marks the lowest test risk and bold the tightest bound per model and loss.}
\label{tab:pipeline-beavertails}
\setlength{\tabcolsep}{4pt}
\renewcommand{\arraystretch}{1.10}
\resizebox{\textwidth}{!}{%
\begin{tabular}{l rrr @{\hspace{12pt}} rrr @{\hspace{12pt}} rrr @{\hspace{12pt}} rrr @{\hspace{12pt}} rrr}
\toprule
 & \multicolumn{3}{c}{\shortstack{Qwen3Guard\\0.6B}} & \multicolumn{3}{c}{\shortstack{Llama-Guard-3\\1B}} & \multicolumn{3}{c}{\shortstack{Qwen3Guard\\4B}} & \multicolumn{3}{c}{\shortstack{Llama-Guard-3\\8B}} & \multicolumn{3}{c}{\shortstack{Qwen3Guard\\8B}} \\
\cmidrule(lr){2-4}\cmidrule(lr){5-7}\cmidrule(lr){8-10}\cmidrule(lr){11-13}\cmidrule(lr){14-16}
\textit{0--1 loss} & FT & Q & Bound & FT & Q & Bound & FT & Q & Bound & FT & Q & Bound & FT & Q & Bound \\
\cmidrule(lr){1-16}
LoRA & \underline{.146} & .146 & 1.000 & \underline{.146} & .146 & 1.000 & \underline{.141} & .141 & 1.000 & .141 & \underline{.140} & 1.000 & .142 & \underline{.141} & 1.000 \\
 & \multicolumn{3}{c}{\scriptsize(r=4, 4-bit)} & \multicolumn{3}{c}{\scriptsize(r=4, 4-bit)} & \multicolumn{3}{c}{\scriptsize(r=4, 4-bit)} & \multicolumn{3}{c}{\scriptsize(r=2, 4-bit)} & \multicolumn{3}{c}{\scriptsize(r=4, 4-bit)} \\
\addlinespace[3pt]
LoRA-XS & .161 & .163 & .213 & .174 & .183 & .243 & .160 & .160 & .230 & .182 & .180 & .240 & .160 & .168 & .225 \\
 & \multicolumn{3}{c}{\scriptsize(r=2, 2-bit)} & \multicolumn{3}{c}{\scriptsize(r=4, 2-bit)} & \multicolumn{3}{c}{\scriptsize(r=2, 4-bit)} & \multicolumn{3}{c}{\scriptsize(r=2, 2-bit)} & \multicolumn{3}{c}{\scriptsize(r=2, 2-bit)} \\
\addlinespace[3pt]
LoRA-GT & .166 & .168 & \textbf{.174} & .213 & .213 & \textbf{.226} & .165 & .165 & \textbf{.174} & .180 & .182 & \textbf{.200} & .163 & .162 & \textbf{.169} \\
 & \multicolumn{3}{c}{\scriptsize(u=1, 4-bit)} & \multicolumn{3}{c}{\scriptsize(u=4, 4-bit)} & \multicolumn{3}{c}{\scriptsize(u=1, 4-bit)} & \multicolumn{3}{c}{\scriptsize(u=4, 4-bit)} & \multicolumn{3}{c}{\scriptsize(u=1, 2-bit)} \\
\midrule
\textit{Brier loss} & FT & Q & Bound & FT & Q & Bound & FT & Q & Bound & FT & Q & Bound & FT & Q & Bound \\
\cmidrule(lr){1-16}
LoRA & \underline{.103} & .103 & 1.000 & \underline{.103} & .103 & 1.000 & \underline{.100} & .100 & 1.000 & .101 & \underline{.100} & 1.000 & \underline{.100} & .100 & 1.000 \\
 & \multicolumn{3}{c}{\scriptsize(r=4, 4-bit)} & \multicolumn{3}{c}{\scriptsize(r=4, 4-bit)} & \multicolumn{3}{c}{\scriptsize(r=4, 4-bit)} & \multicolumn{3}{c}{\scriptsize(r=2, 4-bit)} & \multicolumn{3}{c}{\scriptsize(r=4, 4-bit)} \\
\addlinespace[3pt]
LoRA-XS & .115 & .117 & .162 & .136 & .137 & .182 & .111 & .117 & .192 & .127 & .129 & .180 & .125 & .127 & .192 \\
 & \multicolumn{3}{c}{\scriptsize(r=2, 2-bit)} & \multicolumn{3}{c}{\scriptsize(r=2, 4-bit)} & \multicolumn{3}{c}{\scriptsize(r=4, 2-bit)} & \multicolumn{3}{c}{\scriptsize(r=2, 2-bit)} & \multicolumn{3}{c}{\scriptsize(r=2, 4-bit)} \\
\addlinespace[3pt]
LoRA-GT & .121 & .122 & \textbf{.129} & .147 & .147 & \textbf{.160} & .132 & .129 & \textbf{.146} & .128 & .128 & \textbf{.145} & .134 & .131 & \textbf{.138} \\
 & \multicolumn{3}{c}{\scriptsize(u=1, 4-bit)} & \multicolumn{3}{c}{\scriptsize(u=4, 4-bit)} & \multicolumn{3}{c}{\scriptsize(u=4, 4-bit)} & \multicolumn{3}{c}{\scriptsize(u=4, 4-bit)} & \multicolumn{3}{c}{\scriptsize(u=1, 2-bit)} \\
\bottomrule
\end{tabular}
}
\end{table*}

\FloatBarrier
\subsection{Complete comparison of LoRA-T variants}
\label{app:tinylora-variants}

\begin{figure}[ht!]
    \centering
    \includegraphics[width=1.0\linewidth]{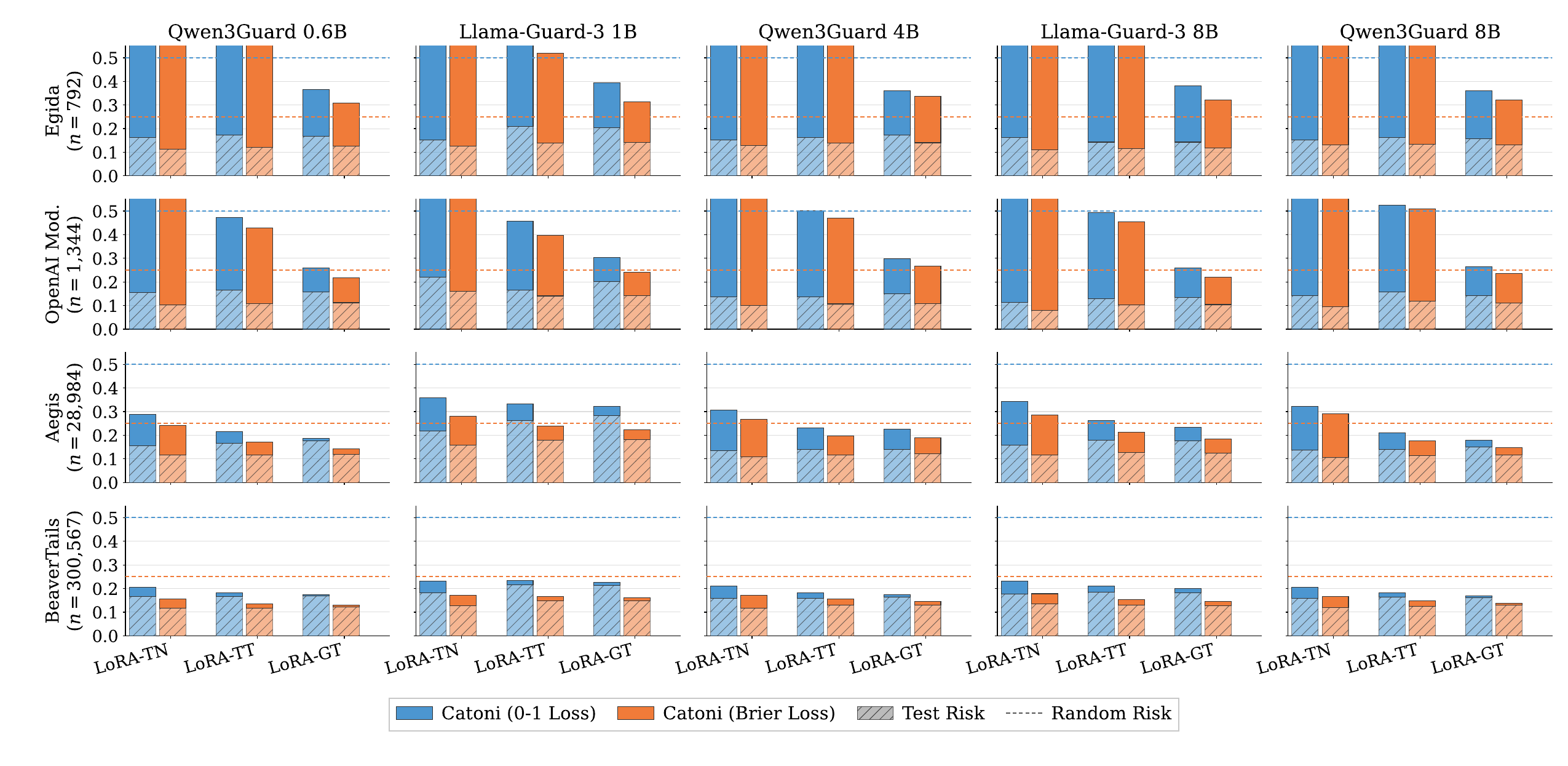}
    \caption{\textbf{Complete comparison of LoRA-T variants.} Catoni PAC-Bayes certificates under $0$--$1$ and Brier loss across all five models and four benchmarks. Hatching shows test risk and dashed lines mark random-risk baselines.}
    \label{fig:tinylora-variants}
\end{figure}

\Cref{fig:tinylora-variants} extends the Llama Guard 3 1B and 8B comparison in \Cref{fig:compression}(b) to all five models. The same pattern holds across all four benchmarks, with LoRA-GT giving the tightest certificates under both losses and generally modest changes in test risk.

\FloatBarrier
\subsection{Transfer to a third model family}
\label{app:nemotron-results}

\begin{figure}[ht!]
    \centering
    \includegraphics[width=0.8\linewidth]{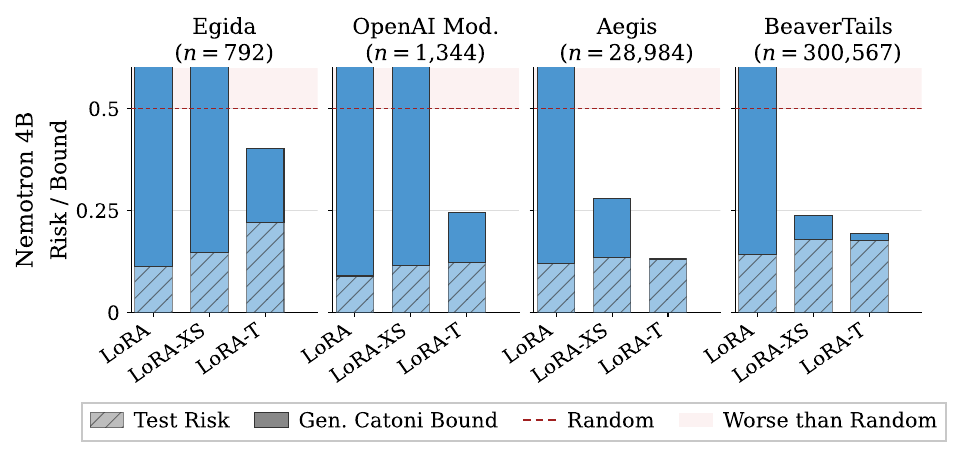}
    \caption{\textbf{Certification of Nemotron 4B under $0$--$1$ loss.} Test risk (hatched) and PAC-Bayes certificate across four benchmarks. LoRA-T yields the tightest certificate on each benchmark. The dashed line marks risk $0.5$.}
    \label{fig:nemotron-results}
\end{figure}

We also evaluate Nemotron 4B under $0$--$1$ loss on all four benchmarks (\Cref{fig:nemotron-results}) to test if our analyses extend beyond the two model families we have focussed on in the main section of this work. We again find that LoRA-T gives the tightest certificate on each benchmark, with substantially looser bounds for LoRA and LoRA-XS. The advantage of compact adaptations therefore extends to a third model family beyond Llama Guard and Qwen3Guard.

\FloatBarrier
\subsection{What sets description length}
\FloatBarrier
\subsubsection{Adaptor description lengths}
\label{sec:desc-lengths}

\begin{figure}[ht!]
    \centering
    \includegraphics[width=1.0\linewidth]{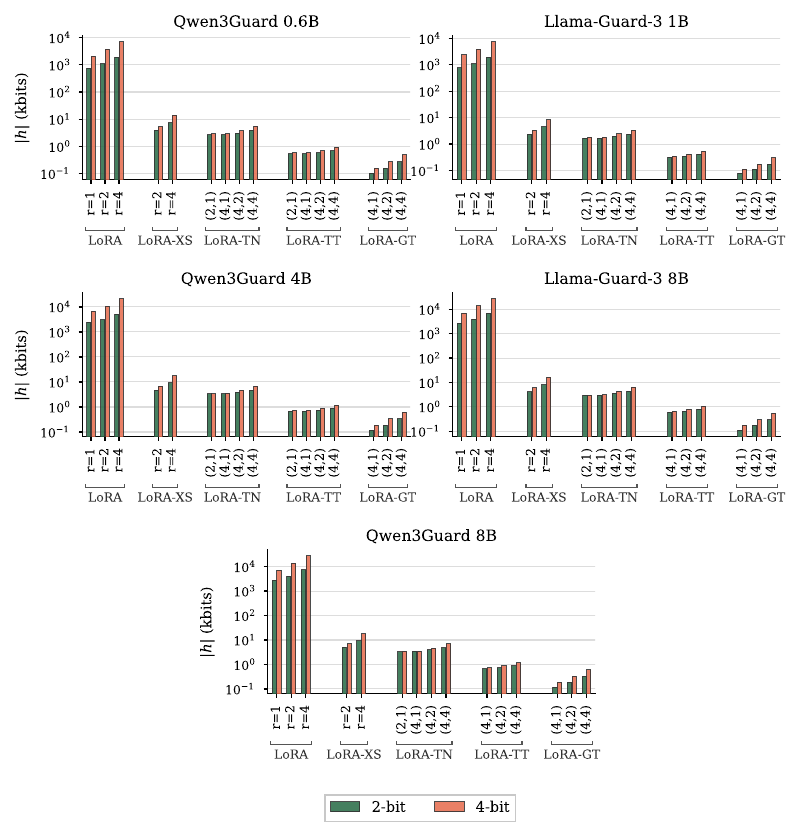}
\caption{\textbf{Adaptor description lengths.} Compressed description lengths $|h|$ (log scale) for all adaptor families and quantisation settings, at $2$ and $4$ bits. LoRA-T settings are given as $(r,u)$. Across guardrail models, LoRA, LoRA-XS, LoRA-TN, LoRA-TT and LoRA-GT exhibit progressively shorter codes, with tying and global-scale quantisation yielding the smallest description lengths.
}
    \label{fig:desc_lengths}
\end{figure}

\Cref{fig:desc_lengths} reports the compressed description lengths $|h|$ for every adaptor family and quantisation setting considered. Across all guardrail models the ordering is consistent. LoRA is largest by three to four orders of magnitude, followed by LoRA-XS, LoRA-TN, LoRA-TT, and finally LoRA-GT, with tying and global-scale quantisation each contributing a further reduction. \Cref{fig:compression} shows that this ordering maps directly onto certificate tightness, with LoRA-GT giving the tightest bounds with a modest increase in test risk.

\FloatBarrier
\subsubsection{Coding scheme comparison}
\label{sec:coding-comparison}

\Cref{tab:elias-comparison} compares hypothesis description lengths under Elias-$\gamma$ and Elias-$\delta$ coding of the integer $|h|$, for Llama-Guard-3 1B at $2$-bit quantisation. Elias $\delta$ is shorter for every adaptor, by $4$ to $12$ bits. This is small relative to $|h|$ itself, but both are valid prefix-free codes, so the tighter one can be substituted with no change to the analysis. We use Elias $\delta$ in all reported bounds, and the same substitution could be applied to prior compression-based certificates \citep{lotfi2024non}.

\begin{table}[h]
\centering
\caption{\textbf{Coding scheme ablation, Llama-Guard-3 1B} (BeaverTails, $b\!=\!2$). Elias-$\delta$ against Elias-$\gamma$ coding of the hypothesis description length (bits).}
\label{tab:elias-comparison}
\small
\begin{tabular}{l r r r}
\toprule
Adaptor & \multicolumn{1}{c}{Elias-$\delta$} & \multicolumn{1}{c}{Elias-$\gamma$} & \multicolumn{1}{c}{Difference} \\
\midrule
LoRA ($r\!=\!4$) & 1,974,975 & 1,974,987 & +12 \\
LoRA-XS ($r\!=\!2$) & 2,588 & 2,593 & +5 \\
LoRA-XS ($r\!=\!4$) & 4,501 & 4,507 & +6 \\
LoRA-TN ($r\!=\!4,\, u\!=\!1$) & 1,691 & 1,695 & +4 \\
LoRA-TN ($r\!=\!4,\, u\!=\!2$) & 2,108 & 2,113 & +5 \\
\bottomrule
\end{tabular}
\end{table}

\FloatBarrier
\subsubsection{Compressor choice: Brotli versus LZMA and gzip}
\label{app:dl-compressor}

The final component of the description length is the lossless compressor applied to the encoded bit-stream (\Cref{app:certconstruct}). \Cref{tab:compressors} compares Brotli (\texttt{quality=11}) against LZMA (\texttt{preset=9}) \citep{python_lzma_docs} and gzip (\texttt{compresslevel=9}) \citep{python_gzip_docs}, the latter being the standard choice in prior compression-based PAC-Bayes work on language models \citep{lotfi2024unlocking}. Brotli produces the shortest compressed output in every row of \Cref{tab:compressors}, between $5$ and $17\%$ shorter than gzip. The complexity penalties, and hence the certificates, of prior work could therefore be tightened by substituting a stronger compressor, with no change to the analysis.

\begin{table}[ht!]
\centering
\caption{\textbf{Compression algorithm comparison.} Compressed output size (in bits) under three lossless compressors for LoRA-based adaptor configurations on Llama-Guard-3 1B (BeaverTails, 2-bit quantisation). The raw column is given in bytes. Brotli is shortest in every row.}
\label{tab:compressors}
\begin{tabular}{lrrrr}
\toprule
Adaptor & Raw (B) & LZMA-9 & gzip-9 & Brotli-11 \\
\midrule
LoRA-TN ($r\!=\!4,\, u\!=\!1$) & 252 & 2,304 & 1,992 & \textbf{1,672} \\
LoRA-TN ($r\!=\!4,\, u\!=\!4$) & 336 & 3,072 & 2,696 & \textbf{2,336} \\
LoRA-XS ($r\!=\!2$) & 336 & 3,072 & 2,712 & \textbf{2,568} \\
LoRA-XS ($r\!=\!4$) & 672 & 5,568 & 5,032 & \textbf{4,480} \\
LoRA ($r\!=\!1$) & 176,576 & 830,752 & 868,672 & \textbf{810,384} \\
LoRA ($r\!=\!4$) & 704,960 & 2,036,640 & 2,371,144 & \textbf{1,974,944} \\
\bottomrule
\end{tabular}
\end{table}

\FloatBarrier
\subsection{Effect of the quantisation scale}
\label{app:scale-intervention}

We compare the no-clipping scale $s^*$ from \Cref{prop:scale} with the searched scale from \Cref{sec:global-scale}, holding each adaptor and bit-width fixed so that only $s$ changes. The searched scale generally tightens the bound by reducing certification-sample distortion $\delta_n$. At $3$ and $4$ bits, description length is unchanged, so this reduction directly accounts for the tighter bounds (\Cref{tab:scale-intervention}). The effect on test risk is clearest at $2$ bits and is small at $4$ bits, where $\delta_n \approx 0$ under both scales.

\begin{table}[h]
\centering
\caption{\textbf{Effect of scale search on LoRA-GT.} Mean distortion is lower with the searched scale at every bit-width. Counts report changes relative to the no-clipping scale $s^*$. $\Delta|h|$ is the mean change in description length.}
\label{tab:scale-intervention}
\small
\setlength{\tabcolsep}{4pt}
\begin{tabular}{crrrrrrrrrr}
\toprule
$b$ & Adaptors & \multicolumn{2}{c}{Mean distortion $\delta_n$}
& \multicolumn{3}{c}{Test risk}
& \multicolumn{3}{c}{Certificate} & $\Delta|h|$ \\
\cmidrule(lr){3-4}\cmidrule(lr){5-7}\cmidrule(lr){8-10}
& & $s^*$ & Searched & Lower & Higher & Equal & Tighter & Looser & Equal & (bits) \\
\midrule
$2$ & $60$ & $0.125$ & $\mathbf{0.065}$ & $42$ & $18$ & $0$ & $\mathbf{40}$ & $19$ & $1$ & $+3.6$ \\
$3$ & $48$ & $0.037$ & $\mathbf{0.024}$ & $26$ & $19$ & $3$ & $\mathbf{33}$ & $13$ & $2$ & $0$ \\
$4$ & $60$ & $0.006$ & $\mathbf{0.004}$ & $26$ & $26$ & $8$ & $\mathbf{31}$ & $24$ & $5$ & $0$ \\
\bottomrule
\end{tabular}
\end{table}

\FloatBarrier
\subsection{Comparison to validation-based bounds}
\label{app:validation-comparison}
Validation bounds offer a simple alternative to PAC-Bayes certification, but require held-out data and do not account for adaptor complexity. We examine how compression and the fraction of training samples reserved for validation affect these guarantees, then compare selecting adaptors by their certificate or validation risk.

\FloatBarrier
\subsubsection{Validation-based generalisation bounds}
 Let $S_{\mathrm{val}} = \{(x_i, y_i)\}_{i=1}^{n_{\mathrm{val}}} \sim (\mathbb{P}_{\mathcal{D},G})^{n_{\mathrm{val}}}$ denote an i.i.d.\ validation sample, and let $\widehat{\mathcal{R}}_{S_{\mathrm{val}}}(h)$ denote the empirical risk of a hypothesis $h \in \mathcal{H}_J$, fixed independently of $S_{\mathrm{val}}$, on it. Hoeffding's inequality \citep{hoeffding1963probability} yields guarantees of the form
\begin{equation}
\mathcal{R}_G(h) \le \widehat{\mathcal{R}}_{S_{\mathrm{val}}}(h) + \sqrt{\frac{\log(1/\beta)}{2 n_{\mathrm{val}}}},
\end{equation}
which hold with probability at least $1 - \beta$ over the draw of $S_{\mathrm{val}}$. An often tighter alternative for losses in $[0,1]$ is the KL-based concentration inequality for Bernoulli random variables \citep{seeger2002pac}. With probability at least $1-\beta$, either $\mathcal{R}_G(h) \le \widehat{\mathcal{R}}_{S_{\mathrm{val}}}(h)$ or
\begin{equation}
\mathrm{kl}\big(\widehat{\mathcal{R}}_{S_{\mathrm{val}}}(h) \,\big\|\, \mathcal{R}_G(h)\big)
\;\le\;
\frac{\log(1/\beta)}{n_{\mathrm{val}}},
\end{equation}
where $\mathrm{kl}$ is the binary Kullback--Leibler divergence of \Cref{sec:alt_bounds}. Inverting this inequality on its upper branch as described by \citet{dziugaite2017computing} gives an upper bound on the risk.

These guarantees answer a different question from ours in two respects. First, they do not involve the training procedure or hypothesis class. They certify a fixed model on unseen data, and for a fixed validation set they depend only on empirical validation error and sample size, so they do not track differences in hypothesis complexity or structural effects such as compressibility. Second, they require a held-out split, obtained with care to avoid leakage. This is often taken as a split from the training set of examples, which reduces the effective training set, a problem within the small-data regimes often seen in safety datasets.

\FloatBarrier
\subsubsection{Validation bounds track empirical risk}

To illustrate the limitations of validation-based guarantees, we examine the behaviour of KL-inversion bounds across the models and datasets of \Cref{sec:experiments}.

\begin{figure}[ht!]
    \centering
    \includegraphics[width=1.0\linewidth]{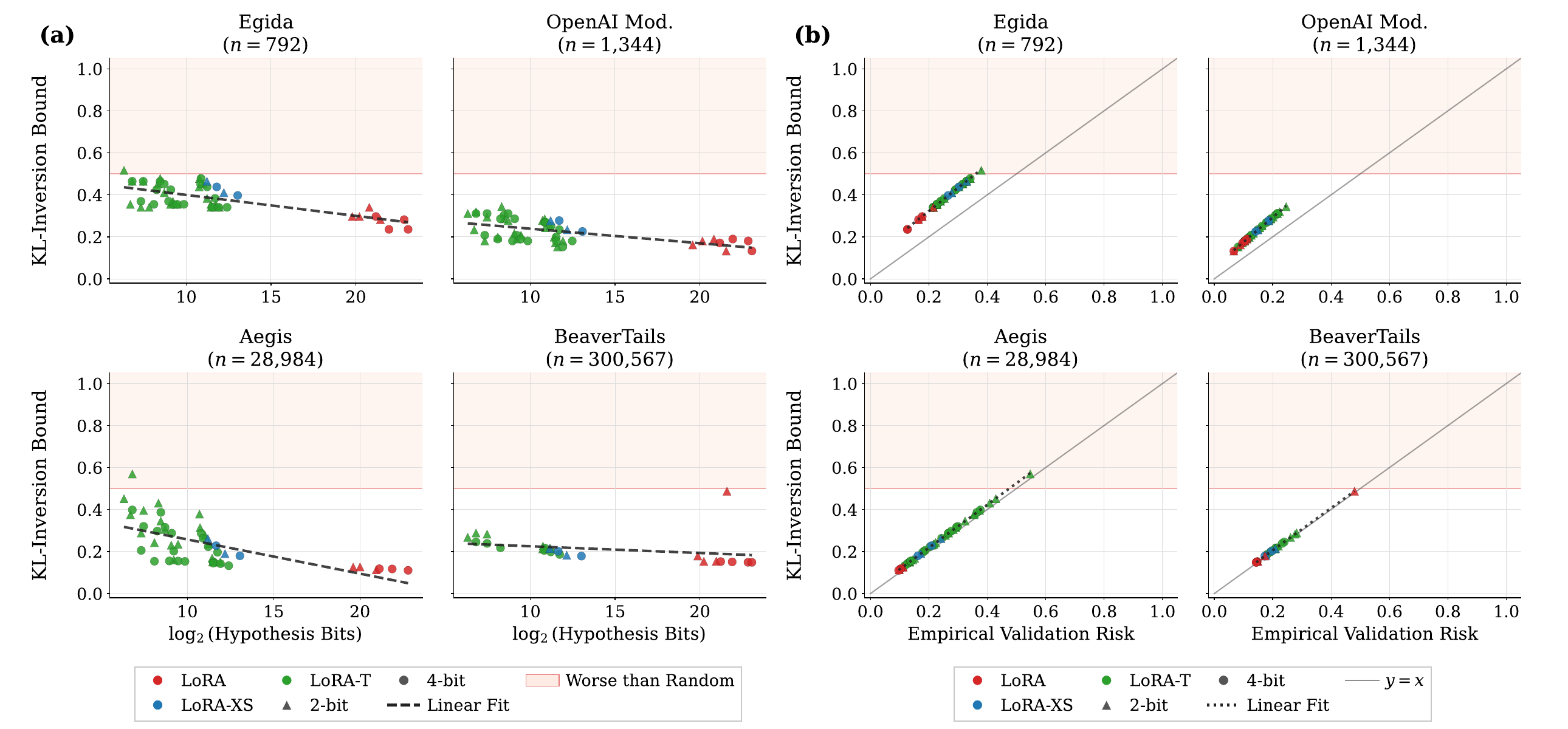}
    \caption{\textbf{Validation-based KL-inversion bound trends.} \textbf{(a)} Bound against description length, which does not enter the validation-bound formula. \textbf{(b)} Bound against empirical validation risk is near-linear, since the bound is a deterministic function of validation risk and $n_{\mathrm{val}}$. This contrasts with the PAC-Bayes certificates of \Cref{fig:compression}, which vary systematically with compressibility.}
    \label{fig:val_bounds}
\end{figure}

\Cref{fig:val_bounds}(a) plots the KL-inversion bound against hypothesis description length. Unlike the PAC-Bayes certificates of \Cref{fig:compression}, the validation bound has no explicit dependence on $|h|$. Adaptors differing by three orders of magnitude in description length receive similar bounds whenever their empirical performance is comparable. \Cref{fig:val_bounds}(b) confirms why. For fixed $n_{\mathrm{val}}$ and $\beta$ the bound is a deterministic function of $\widehat{\mathcal{R}}_{S_{\mathrm{val}}}(h)$ alone, so plotting the two against one another recovers that function exactly. The bound rescales empirical performance and carries no further information about the hypothesis.

A validation bound gives a $10^3$-bit adaptor and a $10^6$-bit adaptor with equal held-out error the same guarantee, because complexity does not enter it. It therefore cannot capture the systematic dependence of generalisation on compressibility reported in prior work \citep{lotfi2022pac, lotfi2024non} and in our experiments.

The PAC-Bayes certificates in the main text behave differently. As \Cref{fig:compression} shows, they vary systematically with hypothesis description length, with more compressed adaptations yielding consistently tighter guarantees, because the explicit complexity term distinguishes hypotheses beyond their empirical fit. PAC-Bayes characterises generalisation through a trade-off between empirical risk and complexity, so tight certificates arise when a hypothesis balances the two. In our experiments, short-description-length adaptations keep strong performance at a much smaller complexity penalty, giving tighter certificates.

\FloatBarrier
\subsubsection{Varying the validation fraction}

We vary the fraction of training data held out for validation from $0.1\%$ to $50\%$ and compare KL-inversion bounds with PAC-Bayes certificates computed on the full training set. Holding out more data reduces the concentration penalty but leaves less data for adaptation.

\begin{figure}[ht!]
    \centering
    \includegraphics[width=0.6\linewidth]{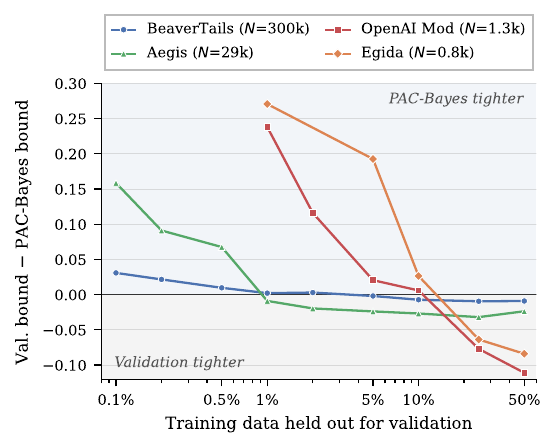}
    \vspace{-0.5em}
    \caption{\textbf{Validation bounds against PAC-Bayes certificates across held-out fractions.} Validation-based KL-inversion bound minus the PAC-Bayes certificate. Positive values indicate a tighter PAC-Bayes certificate, obtained without holding out data.}
    \label{fig:validation-sweep}
    \vspace{-0.5em}
\end{figure}

PAC-Bayes gives tighter certificates at small held-out fractions on every benchmark (\Cref{fig:validation-sweep}), by up to $0.27$ on Egida and $0.24$ on OpenAI Moderation. Validation bounds become tighter at around $10\%$ held out on these two benchmarks and $1\%$ on Aegis. On BeaverTails, the bounds remain within $0.03$ of each other. PAC-Bayes therefore offers the largest benefit on the smaller benchmarks, where competitive validation bounds require holding out a substantial share of the data.

Overall, PAC-Bayes avoids the trade-off between using scarce training samples for adaptation and reserving them for validation-based certification.

\clearpage
\subsubsection{Selecting adaptors by certificate or validation risk}
\label{app:selection-vs-validation}

\Cref{tab:selection-vs-validation} compares two rules for selecting a compressed adaptor, the tightest PAC-Bayes certificate, as in \Cref{fig:main-results}, or the lowest empirical risk on a $10\%$ held-out validation split. Validation-selected models are trained on $90\%$ of the training set with the remaining $10\%$ held out for selection. Certificates for both selection rules are computed on the full training set of size $n$.

Average test risk is similar under the two rules, $0.217$ for certificate selection and $0.209$ for validation selection, while certificate selection gives tighter guarantees on every benchmark. The largest gains are $0.17$ on Egida and $0.07$ on OpenAI Moderation, where validation selects longer $u=4$ adaptors with larger complexity penalties at small $n$.

\begin{table}[h]
\centering
\vspace{-0.5em}
\caption{\textbf{Selecting LoRA-GT configurations with and without a validation split.} Llama-Guard-3 1B over $u \in \{1,2,4\}$ and $b \in \{2,3,4\}$, with $\lceil\log_2 9\rceil = 4$ selection bits charged to every bound. Entries give the selected $(u,b)$, its test risk and its certificate. Best Test is the lowest test risk over the grid.}
\label{tab:selection-vs-validation}
\vspace{-0.5em}
\small
\setlength{\tabcolsep}{4pt}
\begin{tabular}{lrccccccc}
\toprule
 & & \multicolumn{3}{c}{Certificate} & \multicolumn{3}{c}{Validation} & \\
\cmidrule(lr){3-5} \cmidrule(lr){6-8}
Dataset & $n$ & $(u,b)$ & Test & Bound & $(u,b)$ & Test & Bound & Best Test \\
\midrule
Egida & 792 & (1, 2) & .204 & \textbf{.394} & (4, 4) & .168 & .568 & .153 \\
OpenAI Moderation & 1,344 & (1, 3) & .164 & \textbf{.297} & (4, 3) & .170 & .366 & .164 \\
Aegis & 28,984 & (4, 3) & .287 & \textbf{.316} & (4, 4) & .284 & .326 & .268 \\
BeaverTails & 300,567 & (4, 4) & .213 & \textbf{.226} & (4, 3) & .215 & .227 & .213 \\
\bottomrule
\end{tabular}
\end{table}

\clearpage
\normalsize

\section*{NeurIPS Paper Checklist}

\begin{enumerate}

\item {\bf Claims}
    \item[] Question: Do the main claims made in the abstract and introduction accurately reflect the paper's contributions and scope?
    \item[] Answer: \answerYes{}
    \item[] Justification: Claims in the abstract and introduction match theoretical and empirical results (Sections 1 and 5).

\item {\bf Limitations}
    \item[] Question: Does the paper discuss the limitations of the work performed by the authors?
    \item[] Answer: \answerYes{}
    \item[] Justification: Limitations, including assumptions and scope, are discussed in the Discussion (Section 6).

\item {\bf Theory assumptions and proofs}
    \item[] Question: For each theoretical result, does the paper provide the full set of assumptions and a complete (and correct) proof?
    \item[] Answer: \answerYes{}
    \item[] Justification: All assumptions are stated and proofs are provided in the appendix (Section 4, Appendix A).

    \item {\bf Experimental result reproducibility}
    \item[] Question: Does the paper fully disclose all the information needed to reproduce the main experimental results of the paper to the extent that it affects the main claims and/or conclusions of the paper (regardless of whether the code and data are provided or not)?
    \item[] Answer: \answerYes{}
    \item[] Justification: Experimental setup, datasets, and methods are described in sufficient detail (Section 5, Appendices). We will also release open source code to support the paper.

\item {\bf Open access to data and code}
    \item[] Question: Does the paper provide open access to the data and code, with sufficient instructions to faithfully reproduce the main experimental results, as described in supplemental material?
    \item[] Answer: \answerYes{}
    \item[] Justification: Code is linked in the paper.

\item {\bf Experimental setting/details}
    \item[] Question: Does the paper specify all the training and test details (e.g., data splits, hyperparameters, how they were chosen, type of optimizer) necessary to understand the results?
    \item[] Answer: \answerYes{}
    \item[] Justification: Training, datasets, and hyperparameters are described in the main text and appendices.

\item {\bf Experiment statistical significance}
    \item[] Question: Does the paper report error bars suitably and correctly defined or other appropriate information about the statistical significance of the experiments?
    \item[] Answer: \answerYes{}
    \item[] Justification: We show variability via bootstrap confidence intervals in Figure 3.

\item {\bf Experiments compute resources}
    \item[] Question: For each experiment, does the paper provide sufficient information on the computer resources (type of compute workers, memory, time of execution) needed to reproduce the experiments?
    \item[] Answer: \answerYes{}
    \item[] Justification: Compute details are explicitly reported in \Cref{sec:compute_resources}.
    
\item {\bf Code of ethics}
    \item[] Question: Does the research conducted in the paper conform, in every respect, with the NeurIPS Code of Ethics \url{https://neurips.cc/public/EthicsGuidelines}?
    \item[] Answer: \answerYes{}
    \item[] Justification: The work complies with the NeurIPS Code of Ethics and uses public datasets.

\item {\bf Broader impacts}
    \item[] Question: Does the paper discuss both potential positive societal impacts and negative societal impacts of the work performed?
    \item[] Answer: \answerYes{}
    \item[] Justification: Impacts on reliability and safety are discussed (Section 6).
    
\item {\bf Safeguards}
    \item[] Question: Does the paper describe safeguards that have been put in place for responsible release of data or models that have a high risk for misuse (e.g., pre-trained language models, image generators, or scraped datasets)?
    \item[] Answer: \answerNA{}
    \item[] Justification: No new models or datasets are released.

\item {\bf Licenses for existing assets}
    \item[] Question: Are the creators or original owners of assets (e.g., code, data, models), used in the paper, properly credited and are the license and terms of use explicitly mentioned and properly respected?
    \item[] Answer: \answerYes{}
    \item[] Justification: All datasets and models are cited and publicly available.

\item {\bf New assets}
    \item[] Question: Are new assets introduced in the paper well documented and is the documentation provided alongside the assets?
    \item[] Answer: \answerNA{}
    \item[] Justification: No new datasets, models, or code are introduced.

\item {\bf Crowdsourcing and research with human subjects}
    \item[] Question: For crowdsourcing experiments and research with human subjects, does the paper include the full text of instructions given to participants and screenshots, if applicable, as well as details about compensation (if any)? 
    \item[] Answer: \answerNA{}
    \item[] Justification: No human subjects or crowdsourcing are involved.

\item {\bf Institutional review board (IRB) approvals or equivalent for research with human subjects}
    \item[] Question: Does the paper describe potential risks incurred by study participants, whether such risks were disclosed to the subjects, and whether Institutional Review Board (IRB) approvals (or an equivalent approval/review based on the requirements of your country or institution) were obtained?
    \item[] Answer: \answerNA{}
    \item[] Justification: No human subjects research is conducted.

\item {\bf Declaration of LLM usage}
    \item[] Question: Does the paper describe the usage of LLMs if it is an important, original, or non-standard component of the core methods in this research? Note that if the LLM is used only for writing, editing, or formatting purposes and does \emph{not} impact the core methodology, scientific rigor, or originality of the research, declaration is not required.
    \item[] Answer: \answerNA{}
    \item[] Justification: LLMs are evaluated, not used as part of the proposed method.

\end{enumerate}

\end{document}